\documentclass[letterpaper]{article} 
\usepackage{multirow}
\usepackage{amsmath}
\usepackage{booktabs} 
\usepackage{tabularx}
\usepackage{amssymb}
\usepackage{pifont}
\newcommand{\cmark}{\ding{52}}%
\newcommand{\xmark}{\ding{56}}%
\usepackage{makecell}
\usepackage{aaai25}  
\usepackage{times}  
\usepackage{helvet} 
\usepackage{courier}  
\usepackage[hyphens]{url}  
\usepackage{graphicx} 
\usepackage{natbib}  
\usepackage{caption} 
\usepackage{algorithm}
\usepackage{algorithmic}

\usepackage{newfloat}
\usepackage{listings}
\DeclareCaptionStyle{ruled}{labelfont=normalfont,labelsep=colon,strut=off} 
\floatstyle{ruled}
\newfloat{listing}{tb}{lst}{}
\floatname{listing}{Listing}

\nocopyright 

\title{LLM-RG4: Flexible and Factual Radiology Report Generation across Diverse Input Contexts}
\author{
    Zhuhao Wang\textsuperscript{\rm 1}, Yihua Sun\textsuperscript{\rm 1}, Zihan Li\textsuperscript{\rm 1}, Xuan Yang\textsuperscript{\rm 1}, Fang Chen\textsuperscript{\rm 2,3}, Hongen Liao\textsuperscript{\rm 1,2,3}\thanks{Corresponding author}
}
\affiliations{
    \textsuperscript{\rm 1}School of Biomedical Engineering, Tsinghua University, Beijing, China\\
    \textsuperscript{\rm 2}School of Biomedical Engineering, and \textsuperscript{\rm 3}Institute of Medical Robotics, Shanghai Jiao Tong University, Shanghai, China\\
    wangzhuh23@mails.tsinghua.edu.cn, liao@tsinghua.edu.cn
}

\usepackage{bibentry}

\begin{document}

\maketitle

\begin{abstract}

Drafting radiology reports is a complex task requiring flexibility, where radiologists tail content to available information and particular clinical demands. However, most current radiology report generation (RRG) models are constrained to a fixed task paradigm, such as predicting the full ``finding'' section from a single image, inherently involving a mismatch between inputs and outputs. The trained models lack the flexibility for diverse inputs and could generate harmful, input-agnostic hallucinations. To bridge the gap between current RRG models and the clinical demands in practice, we first develop a data generation pipeline to create a new MIMIC-RG4 dataset, which considers four common radiology report drafting scenarios and has perfectly corresponded input and output. Secondly, we propose a novel large language model (LLM) based RRG framework, namely LLM-RG4, which utilizes LLM's flexible instruction-following capabilities and extensive general knowledge. We further develop an adaptive token fusion module that offers flexibility to handle diverse scenarios with different input combinations, while minimizing the additional computational burden associated with increased input volumes. Besides, we propose a token-level loss weighting strategy to direct the model's attention towards positive and uncertain descriptions. Experimental results demonstrate that LLM-RG4 achieves state-of-the-art performance in both clinical efficiency and natural language generation on the MIMIC-RG4 and MIMIC-CXR datasets. We quantitatively demonstrate that our model has minimal input-agnostic hallucinations, whereas current open-source models commonly suffer from this problem.
\end{abstract}

\begin{links}
     \link{Code}{https://github.com/zh-Wang-Med/LLM-RG4}
\end{links}

\section{Introduction}
The automatic generation of textual descriptions for radiographs has the potential to reduce clinicians' workload, enhance the efficiency of image interpretation, and support informed treatment decisions. Numerous works have concentrated on generating the comprehensive findings section of the report from a single radiology image~\cite{li2018hybrid,chen2020generating,chen2021cross,wang2022medical,wang2023metransformer,yan2024ahive}. However, certain information in the report is uninferable within a single image, resulting in a mismatch between the input and the output, as illustrated in Figure 1(a). Concretely, \citet{nguyen2023pragmatic} classify the information in a report into several key components: positive mentions, negative mentions, prior comparisons, prior procedures, image views, doctor communications, and medical recommendations. Notably, elements such as comparisons, procedures, communication and views 
are uninferable within a single image. Current paradigm amplifies model hallucination, reduces model performance, and lowers clinical acceptance.

\begin{figure}[t]
	\centering
	\includegraphics[width=0.9\columnwidth]{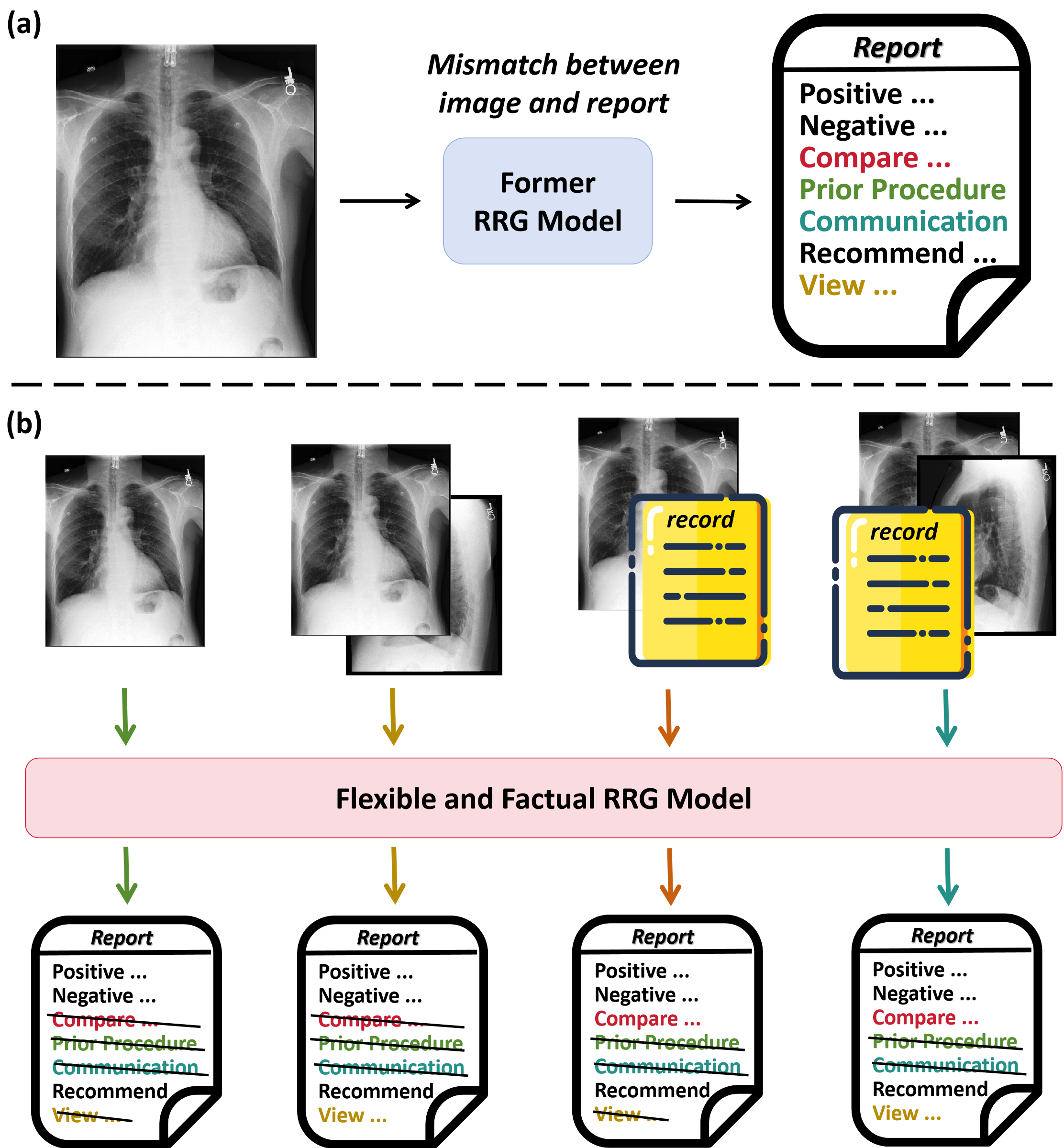} 
	\caption{(a) Mismatch between image and report in typical RRG model. Comparisons, procedures, communication and views are uninferable. (b) A flexible and factual RRG paradigm, which emphasizes the flexibility of input and the alignment between input and output.}
	\label{fig1}
\end{figure}

In response to this phenomenon, several studies have sought to clean and reconstruct the report content. For instance, \citet{ramesh2022improving} introduced the GILBERT model, which utilizes token-by-token classification to eliminate comparative descriptions, \citet{thawakar2024xraygpt} and \citet{nguyen2023pragmatic} leveraged the language comprehension capabilities of large language models to eliminate uninferable descriptions when provided with a single image. However, such approaches drastically reduce the information included in the report, potentially undermining its effectiveness in fulfilling the intended function of radiology reports \cite{hartung2020create}. Furthermore, several studies have explored multi-view modeling~\cite{yuan2019automatic,miura2021improving,lee2023unified}
 and longitudinal historical information modeling~\cite{dalla2023controllable,sanjeev2024tibix} to incorporate addition valid information. However, their performance deteriorates in the absence of additional information, and continues to produce hallucinations

In clinical practice, doctors adaptively draft reports based on available information and clinical requirements \cite{johnson2019mimic}. The radiologist compare the current findings with previous examinations when historical records are accessible. They integrate information from multiple views if provided, and concentrate on findings within a single frontal view if only one is available. Therefore, a more flexible and factual model for RRG should adapt to diverse input scenarios and produce reports inferred from the input within a unified framework, as illustrated in Figure 1(b). Meanwhile, it is crucial to identify definitive or potential lesions across various input scenarios to ensure timely intervention.

Inspired by clinical practices, this work introduces a flexible and factual framework consisting of a new data generation pipeline and a novel model architecture for RRG. The data generation pipeline produces the MIMIC-RG4 dataset from MIMIC-CXR, taking into account four common input scenarios that involve the integration of multi-view and longitudinal data.
The pipeline comprises a BERT-based discriminator, namely DiscBERT, and a generator, Llama3-70B \cite{llama3modelcard}. It utilizes a cyclic generation approach to ensure that the reconstructed reports closely correspond to the input while minimizing information loss. Additionally, DiscBERT, as a byproduct of the pipeline, allows for quantitative input-agnostic information evaluation. In terms of the model architecture, we introduce LLM-RG4, which utilizes LLM's flexible instruction-following capabilities and extensive general knowledge.~\cite{li2023mimic,li2023blip,guo2023point}. To avoid increasing the computational burden with additional input types, we design an adaptive token fusion module that accommodates various inputs while maintaining a consistent token count. The underlying intuition is that an efficient and high-fidelity information encoding for multimodal large language model (MLLM) can be achieved within specialized medical tasks. Furthermore, to improve clinical accuracy, we employ a token-level loss weighting strategy that prioritizes critical diagnoses. This approach enhances clinical efficacy directly at the loss layer, without depending on reinforcement learning \cite{miura2021improving} or classifier-assisted techniques \cite{jin2024promptmrg}. We validate our framework through experiments on MIMIC-CXR and MIMIC-RG4. Our contributions are summarized as follows.

\begin{itemize}
	\item  
         We present a novel paradigm MIMIC-RG4 for pragmatic RRG, introduce a new pipeline for data generation and a product DiscBERT to quantitatively evaluate input-agnostic hallucinations.
	\item We develop LLM-RG4, a LLM-based RRG model that incorporates an adaptive token fusion module to efficiently accommodate different inputs and a token-level loss weighting strategy to enhance diagnostic accuracy.
	\item We conduct extensive experiments demonstrating that LLM-RG4 achieves state-of-the-art performance in CE and NLG dimensions on both the MIMIC-CXR and MIMIC-RG4 datasets, while minimizing input-agnostic hallucinations, thus bridging the gap to clinical practice.
\end{itemize}

\section{MIMIC-RG4 Paradigm}
\subsection{Problem Formulation}
In contrast to the typical report generation paradigm, MIMIC-RG4 exhibits two notable differences. Firstly, it requires the model to be able to handle different input scenarios. Secondly, regardless of the input case, the model generates reports corresponding to the inputs. 

We define four common clinical input scenarios: single view no longitudinal, multi-view no longitudinal, single view with longitudinal, and multi-view with longitudinal. Longitudinal refers to previous X-ray examinations and here we only include previous reports \( T_p \). Single view refers to frontal image \( I_f \). Multi-view refers to frontal and lateral images \( I_l \). Meanwhile, the indication \( T_i \) or history \( T_h \) section is also important for report generation~\cite{miura2021improving,hyland2023maira}, thus we incorporate them as inputs if available. 
Denote the model as \( L_g\) and the current report as  \( T_c\), the entire task is formalized as:
\begin{align}
T_c &= L_g (I_f,I_l,T_p,T_h,T_i) \tag{1} 
\end{align}
where \( I_l \), \( T_p \), \( T_i \), \( T_h \) allow for absence and the corresponding \( T_c \) will vary accordingly. With respect to the contents of reports, we follow the definition of \citet{nguyen2023pragmatic} and reconstruct the report to maximize the retention of effective information. Specifically, for different input scenarios, the communication and prior procedure are removed. Positive mentions, negative mentions, and medical recommendations are retained. View and prior comparison are retained or rewritten depending on the inputs.
\begin{figure}[t]
	\centering
	\includegraphics[width=0.9\columnwidth]{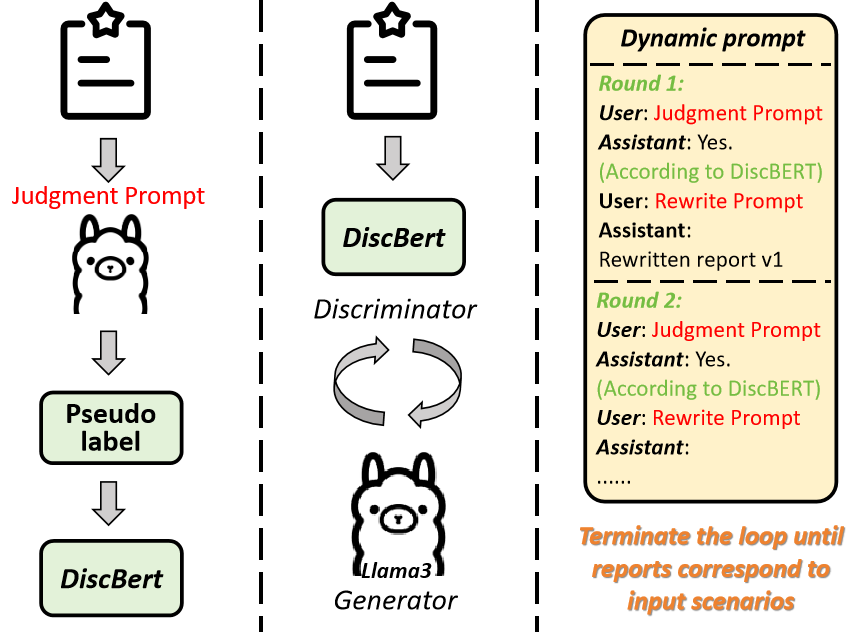} 
	\caption{The pipeline employs an iterative approach that integrates a BERT-based discriminator and a LLM-based generator, ensuring minimal input-agnostic information and effective information loss.}
	\label{fig2}
\end{figure}
\subsection{Dataset Generation Pipeline}
The overall pipeline is shown in Figure 2. Based on problem formulation, the prior comparison, prior procedure, view, and communication sections are of greater importance. 
We utilize Llama3-70B model as the generator to reconstruct reports. However, some challenging cases require multiple modifications to meet the requirements. Therefore, we adopt a cyclic process of judgment, rewriting, and re-evaluation until the report is satisfactory. In iterative modifications, the previous process is also used to form a dynamic prompt that fully leverages the model's chain-of-thought (COT) capability. This is a highly time-consuming process. To expedite it, we train a BERT-based model, DiscBERT, as the discriminator to perform judgment tasks. DiscBERT is trained with the Llama3-70B's judgment results, and exhibits judgement capabilities comparable to Llama3-70B. To preserve the diagnostic information of the report, we employ CheXbert to compute the disease labels for the impression section before and after processing. If there is any change in labels, the corresponding example is discarded. We set the maximum iterations to 3 rounds.

Our integrated approach enables us to leverage the COT of LLM while accelerating the pipeline by reducing the reliance on LLM. Moreover, DiscBERT allows for convenient dataset analysis that distinguishes information categories within the generated reports, offering a tool for evaluating input-agnostic hallucinations. Doctors manually label 200 reports to evaluate the discriminatory performance of DiscBERT and assess the effectiveness of the pipeline. The details about DiscBERT, instructions and evaluation are presented in the supplementary material. 

\begin{figure*}[t]
	\centering
	\includegraphics[width=0.9\textwidth]{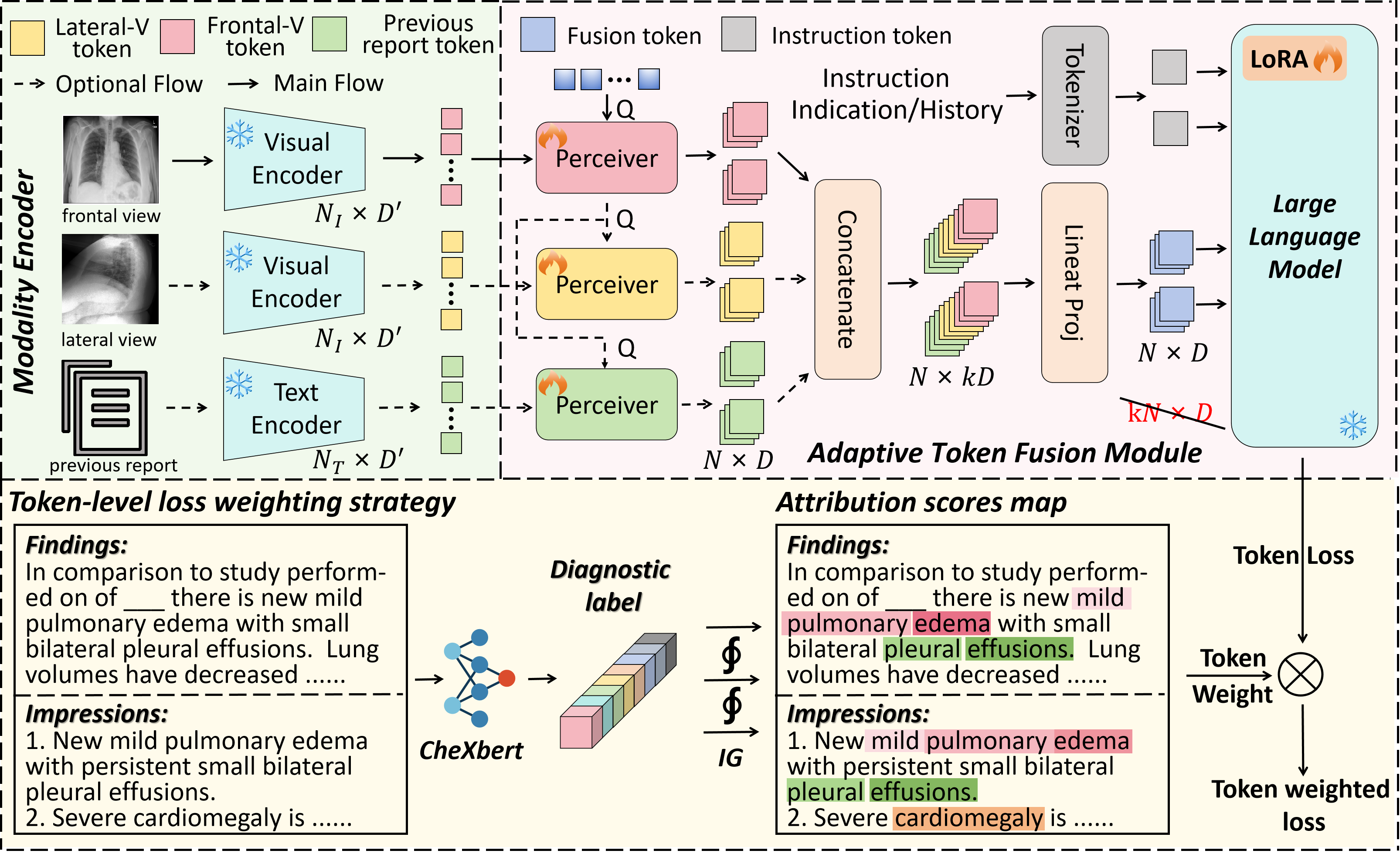}
	\caption{The LLM-RG4 architecture consists of a modality encoder, an adaptive token fusion module, and a token-level loss weighting strategy. The modality encoder extracts features from various modalities. The adaptive token fusion module combines different feature tokens into a fixed length, minimizing computational burden. The token-level loss weighting strategy identifies key diagnoses and adjusts token loss weights, enhancing the model's clinical efficacy across diverse input scenarios.}
	\label{fig3}
\end{figure*}

\subsection{Dataset Statistics}
We utilize the MIMIC-CXR dataset \cite{johnson2019mimic}, which is the only publicly available dataset that encompasses both multi-view and longitudinal information, to generate the MIMIC-RG4 dataset. We evaluate the proportion of reports containing input-agnostic information in the MIMIC-CXR and MIMIC-RG4 datasets under the single view no longitudinal setting, as assessed by DiscBERT. As shown in Table 1, a few cases in MIMIC-CXR meet the criterion of single view no longitudinal. However even in these cases, the reports still include a number of descriptions of prior comparisons and procedures. This indicates that obtaining the dataset for single view no longitudinal data directly from MIMIC-CXR is inappropriate. In contrast, MIMIC-RG4 minimizes the presence of input-agnostic information and features a larger dataset scale. 

\begin{table}[h]
	\centering
        \footnotesize
        \newcolumntype{C}{>{\centering\arraybackslash}X}%
		\begin{tabularx}{\columnwidth}{CCCCCC}
			\toprule
			Dataset	& split	& PC(\%) & PP(\%) & View(\%) & Comm(\%) \\
			\midrule
			\multirow{3}{*}{\makecell{MIMIC\\CXR}}  &Tr/16.9K&39.90&15.99&	1.17&4.56 \\
			& Val/0.1K&38.36&11.28&1.50&3.01 \\
			& Ts/0.1K	&65.63&	25.78&	3.13&	10.16 \\
			\midrule
			\multirow{3}{*}{\makecell{MIMIC\\RG4}} &Tr/172.6K&0.30&0.30&	0.12&0.00\\
			& Val/1.4K&	0.07&	0.07&	0.14&	0.00 \\
            & Ts/2.4K&	0.42&	0.42&	0.04&	0.04 \\
			\bottomrule
		\end{tabularx}
		\caption{Percentage of reports with single image no longitudinal setting, that encompass various categories of information. PC: Prior Comparison; PP: Prior Procedure; Comm: Communication; Tr: train; Ts: test.}
		\label{table1}
	\end{table}

\section{Method}
The overall architecture, depicted in Figure 3, consists of three primary components: modality encoder, adaptive token fusion module (ATF), and token-level loss weighting strategy (TLW). The modality encoder utilizes pretrained encoders to extract features from a range of visual and textual inputs. 
The adaptive token fusion module subsequently compresses and integrates these features into a fixed-length fusion token. This token is then supplied to the LLM alongside instructions and indications/histories for decoding. The token-level loss weighting strategy identifies critical diagnostic tokens and assigns them higher loss weights, thereby directing the model to emphasize positive or ambiguous descriptions regardless of the input conditions.

\begin{table*}[t]
\centering
\footnotesize
\newcolumntype{C}{>{\centering\arraybackslash}X}%
\begin{tabularx}{\linewidth}{lCCCCCCCCCC}
\toprule	 
\multirow{2}{*}{\textbf{Model}} & \multicolumn{3}{c}{\textbf{CE Metrics}} & \multicolumn{3}{c}{\textbf{Clean NLG }} &\multicolumn{3}{c}{\textbf{Original NLG }} & \multirow{2}{*}{\textbf{\textit{hall.}}}\\ 
\cmidrule(r){2-4} \cmidrule(r){5-7} \cmidrule(r){8-10} 
&P & R & F1 & B@1 & B@4 & R-L & B@1 & B@4 & R-L &\\
\midrule
R2Gen\cite{chen2020generating} & 0.456 & 0.306 & 0.366 & 0.363 & 0.090 & 0.269 & 0.356 & 0.097 & 0.267 & 0.779 \\
R2GenCMN\cite{chen2021cross} & 0.486 & 0.400 & 0.439 & 0.385 & 0.102 & 0.278& 0.349  & 0.094 & 0.270 & 0.695 \\
CVT2Dis.\cite{nicolson2023improving} & 0.498 & 0.414 & 0.452 & 0.374 & 0.103 & 0.272 & 0.390 & 0.123 & 0.282 & 0.875 \\
KiUT$^{\dagger}$\cite{huang2023kiut} & 0.371 & 0.318 & 0.321 & - & - & - & 0.391 & 0.113 & 0.285 & - \\
RGRG$^{\dagger}$\cite{tanida2023interactive} & 0.461 & 0.475 & 0.447 & - & - & - & 0.373 & 0.126 & 0.264 & - \\
EKAGen$^{\dagger}$\cite{bu2024instance} & 0.517 & 0.483 & 0.499 & - & - & - & 0.419 & 0.117 & 0.287 & - \\
Promptmrg\cite{jin2024promptmrg} & \textbf{0.618} & 0.491 & 0.548 & 0.326 & 0.080 & 0.261 & 0.381 & 0.096 & 0.258 & 0.896 \\
R2GenGPT(7B)\cite{wang2023r2gengpt} & 0.506 & 0.414 & 0.456 & 0.401 & 0.118 & 0.277 & 0.396 & 0.113 & 0.273 & 0.917 \\
CheXagent(7B)\cite{chen2024chexagent} & 0.506 & 0.306 & 0.381 & 0.265 & 0.058 & 0.239 & 0.189 &0.040 & 0.208 & 0.549 \\
MAIRA-1(7B)$^{\dagger}$\cite{hyland2023maira} & - & - & 0.553 & - & - & - & 0.392 & 0.142 & 0.289 & - \\
Med-PaLM(562B)$^{\dagger}$\cite{tu2024towards}& - & - & 0.516 & - & - & - & 0.317 & 0.115 & 0.275 & - \\
R2-LLM(14.2B)$^{\dagger}$\cite{liu2024bootstrapping} & 0.465 & 0.482 & 0.473 & - & - & - & 0.402 & 0.128 & 0.291 & - \\
InVERGe(7B)$^{\dagger}$\cite{deria2024inverge} & - & - & - & - & - & - & \textbf{0.425} & 0.100 & 0.309 & - \\
\cmidrule(r){1-11}
\textbf{Ours} & 0.583 & \textbf{0.593} & \textbf{0.588} & \textbf{0.498} & \textbf{0.203} & \textbf{0.387} & 0.377 & \textbf{0.144} & \textbf{0.318} & \textbf{0.015} \\
\bottomrule
\end{tabularx}
\caption{\small Comparison with SOTA methods for the setting of \textbf{\textit{sn}}. $\dagger$ indicates the results are quoted from the published literature. \textbf{Clean NLG} refers to using the cleaned reports from MIMIC-RG4 as ground truth, while \textbf{Original NLG} denotes using the original reports from MIMIC-CXR as ground truth. \textbf{\textit{hall.}} means the percentage of reports containing input-agnostic information. The best results are in \textbf{bold}.}
\label{tab:results_sota}
\end{table*}

\subsection{Modality Encoder}
Under different input scenarios, the model can access frontal image \( I_f \), lateral images \( I_l \), and previous examination reports \( T_p \). We utilize frozen image encoder \( E_v \) and text encoder \( E_t \) to obtain corresponding features \( v_f \), \( v_l \), \( v_t \). 
\begin{align}
v_f &= E_v (I_f) \tag{2} \\
v_l &= E_v (I_l) \tag{3} \\
v_t &= E_t (T_p) \tag{4}
\end{align}
where \( v_f \), \( v_l  \in \mathbb{R}^{N_I \times D'} \), \( v_t  \in \mathbb{R}^{N_T \times D'} \), \( D' \) is the dimension of feature obtained from modality encoder, \( N_I \) is the number of visual tokens, \( N_T \) is the number of text tokens. \( N_I \) and \( N_T \) are generally distinct, with \( N_I \) frequently being larger when higher resolution images are employed.

\subsection{Adaptive Token Fusion Module}
Our objective is to maintain a consistent number of feature tokens across different inputs. For instance, with \( I_f \), \( I_l \) and \( T_p \) present, we hope to produce a fused feature with dimensions equivalent to those of \( I_f \) alone. We first employ perceiver \( p_f \), \( p_l \), \( p_t \) \cite{jaegle2021perceiver} and linear layers to further extract and compress modality feature to a consistent dimension \( h_f \), \( h_l \), \( h_t \in \mathbb{R}^{N \times D} \) as follows:
\begin{align}
h_f &= Linear(p_f (v_f,V'))           \tag{5} \\
h_l &= Linear(p_l (v_l,p_f (v_f,V'))) \tag{6} \\
h_t &= Linear(p_t (v_t,p_f (v_f,V'))) \tag{7}
\end{align}
where \( N \) is the compressed number of feature tokens and considerably smaller than \( N_I \), \( D \) is the feature dimension in LLM. Query tokens in \( p_f \) are learnable variables \( V' \in \mathbb{R}^{N \times D'} \), whereas the query tokens in \( p_l \), \( p_t \) are \( p_f (v_f,V'))\) derived from frontal image. This approach leverages the frontal image as the primary feature for modality integration, underscoring its critical role across different scenarios.

Typical visual instruction tuning can be viewed as stacking modality features along the token dimension, and escalates computational complexity with increasing input volumes. Given that instruction tuning enables a LLM to understand unseen visual tokens, it is also feasible to train it to comprehend mixed visual-linguistic tokens similarly. Therefore, we consider compressing information along the feature dimensions of each token. Specifically, we first concatenate \( h_f \), \( h_l \), \( h_t \in \mathbb{R}^{N \times D} \) along the feature dimensions and get \( h_o \in \mathbb{R}^{N \times 3D} \), formulated as:
\begin{align}
h_0 &= concat(h_f,h_l,h_t,dim=1) \tag{8} 
\end{align}
If \( h_l \) or \( h_t \) are not available, we replace it with zeros. To avoid confusion between different modalities, the concatenation order is fixed. A linear projection layer is utilized to maintain the feature dimensions, resulting in the final features \( h'_o \), which are then input into the LLM. 
This approach guarantees that the number of feature tokens remains constant while efficiently encoding modality information across varying input scenarios.

\begin{table*}[t]
\centering
\footnotesize
\newcolumntype{C}{>{\centering\arraybackslash}X}%
\begin{tabularx}{\linewidth}{llCCCCCCCCCC}
\toprule	 
\multirow{2}{*}{\textbf{Dataset}} & \multirow{2}{*}{\textbf{Model}} & \multicolumn{3}{c}{\textbf{CE Metrics}} & \multicolumn{6}{c}{\textbf{NLG Metrics \boldmath$(p < 0.05)$}} & \multirow{2}{*}{\textbf{\textit{hall.}}}\\ 
\cmidrule(r){3-5} \cmidrule(r){6-11}
& &P & R & F1 & B@1 & B@2 & B@3 & B@4 & R-L & MTR &\\
\midrule
\multirow{3}{*}{\textbf{\makecell{RG4\\\textbf{\textit{sn}}}}} & cxrmate$^*$ & 0.572$^{\ddagger}$ & 0.560$^{\ddagger}$
 & 0.566$^{\ddagger}$ & 0.421$^{\ddagger}$ & 0.271$^{\ddagger}$ & 0.179$^{\ddagger}$ & 0.122$^{\ddagger}$ & 0.311$^{\ddagger}$ & 0.174$^{\ddagger}$ & \textbf{0.010} \\
& RadFM & 0.413$^{\ddagger}$ & 0.303$^{\ddagger}$ & 0.350$^{\ddagger}$ & 0.188$^{\ddagger}$ & 0.090$^{\ddagger}$ & 0.048$^{\ddagger}$ & 0.028$^{\ddagger}$ & 0.190$^{\ddagger}$ & 0.094$^{\ddagger}$ & 0.737$^{\ddagger}$ \\
\cmidrule(r){2-12}
& \textbf{Ours} & \textbf{0.588} & \textbf{0.632} & \textbf{0.609} & \textbf{0.479} & \textbf{0.343} & \textbf{0.255} & \textbf{0.196} & \textbf{0.384} &\textbf{0.209} & 0.014 \\
\midrule
\multirow{3}{*}{\textbf{\makecell{RG4\\\textbf{\textit{sw}}}}} & cxrmate$^*$ & 0.573$^{\ddagger}$ & 0.549$^{\ddagger}$ & 0.561$^{\ddagger}$ & 0.361$^{\ddagger}$ & 0.220$^{\ddagger}$ & 0.139$^{\ddagger}$ & 0.093$^{\ddagger}$ & 0.284$^{\ddagger}$ & 0.153$^{\ddagger}$ & \textbf{0.009} \\
& RadFM & 0.508$^{\ddagger}$ & 0.365$^{\ddagger}$ & 0.425$^{\ddagger}$ & 0.211$^{\ddagger}$ & 0.103$^{\ddagger}$ & 0.056$^{\ddagger}$ & 0.033$^{\ddagger}$ & 0.183$^{\ddagger}$ & 0.105$^{\ddagger}$ & 0.092$^{\ddagger}$ \\
\cmidrule(r){2-12}
& \textbf{Ours} & \textbf{0.599} & \textbf{0.622} & \textbf{0.610} & \textbf{0.455} & \textbf{0.321} & \textbf{0.239} & \textbf{0.186} & \textbf{0.382} & \textbf{0.199} & 0.021 \\
\midrule
\multirow{3}{*}{\textbf{\makecell{RG4\\\textbf{\textit{mn}}}}} & cxrmate$^*$ & \textbf{0.544} & 0.522$^{\ddagger}$ & 0.533$^{\ddagger}$ & 0.437$^{\ddagger}$ & 0.289$^{\ddagger}$ & 0.199$^{\ddagger}$ & 0.141$^{\ddagger}$ & 0.332$^{\ddagger}$ & 0.179$^{\ddagger}$ & 0.009 \\
& RadFM & 0.323$^{\ddagger}$ & 0.187$^{\ddagger}$ & 0.237$^{\ddagger}$ & 0.246$^{\ddagger}$ & 0.113$^{\ddagger}$ & 0.060$^{\ddagger}$ & 0.034$^{\ddagger}$ & 0.194$^{\ddagger}$ & 0.104$^{\ddagger}$ & 0.140$^{\ddagger}$ \\
\cmidrule(r){2-12}
& \textbf{Ours} & 0.541 & \textbf{0.578} & \textbf{0.559} & \textbf{0.491} & \textbf{0.359} & \textbf{0.274} & \textbf{0.216} & \textbf{0.405} & \textbf{0.214} &\textbf{0.008} \\
\midrule
\multirow{3}{*}{\textbf{\makecell{RG4\\\textbf{\textit{mw}}}}} & cxrmate$^*$ & 0.548$^{\ddagger}$ & 0.499$^{\ddagger}$ & 0.523$^{\ddagger}$ & 0.379$^{\ddagger}$ & 0.241$^{\ddagger}$ & 0.158$^{\ddagger}$ & 0.110$^{\ddagger}$ & 0.305$^{\ddagger}$ & 0.159$^{\ddagger}$ & 0.007 \\
& RadFM & 0.456$^{\ddagger}$ & 0.297$^{\ddagger}$ & 0.360$^{\ddagger}$ & 0.191$^{\ddagger}$ & 0.095$^{\ddagger}$ & 0.054$^{\ddagger}$ & 0.034$^{\ddagger}$ & 0.178$^{\ddagger}$ & 0.095$^{\ddagger}$ & 0.052$^{\ddagger}$ \\
\cmidrule(r){2-12}
& \textbf{Ours} & \textbf{0.560} & \textbf{0.565} & \textbf{0.563} & \textbf{0.461} & \textbf{0.331} & \textbf{0.250} & \textbf{0.197} & \textbf{0.401} &\textbf{0.204} & \textbf{0.002} \\
\bottomrule
\end{tabularx}
\caption{\small Comparison with SOTA methods supporting MIMIC-RG4 across four settings. $*$ indicates the model is retrained on MIMIC-RG4. $^{\ddagger}$ denotes statistical significance in paired comparisons with LLM-RG4 based on the Wilcoxon signed-rank test. \textbf{\textit{hall.}} means the percentage of reports containing input-agnostic information. The best results are in \textbf{bold}.}
\label{tab:results_sota}
\end{table*}

\subsection{Token-Level Loss Weighting Strategy}
We denote the logit computation function as \( f(\theta) \), where \( \theta \) is the parameter of LLM, denote the current report with a length \( L \) as \( T = [t^1, t^2, t^3, \ldots, t^L] \), and denote the instruction as \( P \). The predicted logit \( o \) is depicted as:
\begin{align}
o^j &= f_\theta (P, h_o, T^{<j})  \tag{9} 
\end{align}
where \( T^{<j} \) represents previous tokens before position \( j \)  in the current report. The loss at token \( t^j \) is depicted as:
\begin{align}
L_{\text{MLE}}^{(t^j)} = -c_j \log_{\text{softmax}} (o^j)  \tag{10} 
\end{align}
where \( c_j=1 \), for \(j=1,2,\ldots,L\), if all tokens are treated equally. To identify key diagnostic tokens in each report and subsequently adjust the coefficient \(c_j\) for each token loss, we utilize CheXbert \cite{smit2020combining} and Integrated Gradients (IG) \cite{sundararajan2017axiomatic}. CheXbert conducts multi-label categorization to identify diagnostic labels \( Y = [y_1, y_2, y_3, \ldots, y_{14}] \) for 14 distinct diseases in reports. \( y_i \in \{-1, 0, 1, 2\} \), \( -1 \) indicates uncertainty, \( 0 \) indicates negative, \( 1 \) indicates positive, \( 2 \) indicates not mentioned.

Given that the 14th category in CheXbert is ``NO FINDING", we focus solely on the attribution maps of the first 13 categories. The detailed algorithm is depicted in Algorithm 1. 
We use CheXbert to identify uncertain or positive diagnostic labels in the report and apply Integrated Gradients to generate attribution maps for each label. We then maximize these maps and smooth them with a Gaussian kernel to mitigate token anomaly. If a token's attribution score exceeds a threshold, the weight of each token in the whole sentence is increased to \(\lambda\). Otherwise, it remains unchanged. Through this approach, we emphasize entire sentences with positive or uncertain diagnoses, minimizing the impact of minor discrepancies between tokens and attribution scores resulting from different tokenizers.

\begin{algorithm}[b!]
\caption{Detailed Procedure of Token Weight \( C \)}
\label{alg:algorithm}
\textbf{Input}: Report \( T = [t^1, t^2, t^3, \ldots, t^L] \), CheXbert \( f_c \)\\
\textbf{Output}: \( C = [c_1, c_2, c_3, \ldots, c_L] \)
\begin{algorithmic}[1] 
\STATE Initialize \( c_i = -1 \)
\STATE Get \( Y = [y_1, y_2, y_3, \ldots, y_{13}] = f_c(T) \)
\STATE \textbf{For} \( y_j \) \textbf{in} \( G \):
\STATE \quad \textbf{if} \( y_j = -1 \) or \( 1 \): \textbf{then}
\STATE \quad \quad \( c_i' = \text{IG}_i(x) \)
\STATE \quad \quad \( c_i = \max(c_i, c_i') \)
\STATE Define \( g_k = \frac{1}{\sqrt{2 \pi \sigma^2}} e^{-\frac{k^2}{2 \sigma^2}} \)
\STATE Split \(C\) into \(M\) sentences \( C^s = [c^1, c^2, c^3, \ldots, c^M] \), \(c^n\) is the \(n\)th sentence's weights with length \(Ln\), \( c^n = [c_1^n, c_2^n, c_3^n, \ldots, c_{L_n}^n] \).
\IF {\( c_i^n > threshold \)}
\STATE \( c^n = \lambda \) and \( \lambda > 1 \)
\ELSE
\STATE \( c^n = 1 \)
\ENDIF
\STATE \textbf{return} \( C \)
\end{algorithmic}
\end{algorithm}

\begin{figure*}[t]
	\centering
	\includegraphics[width=0.9\textwidth]{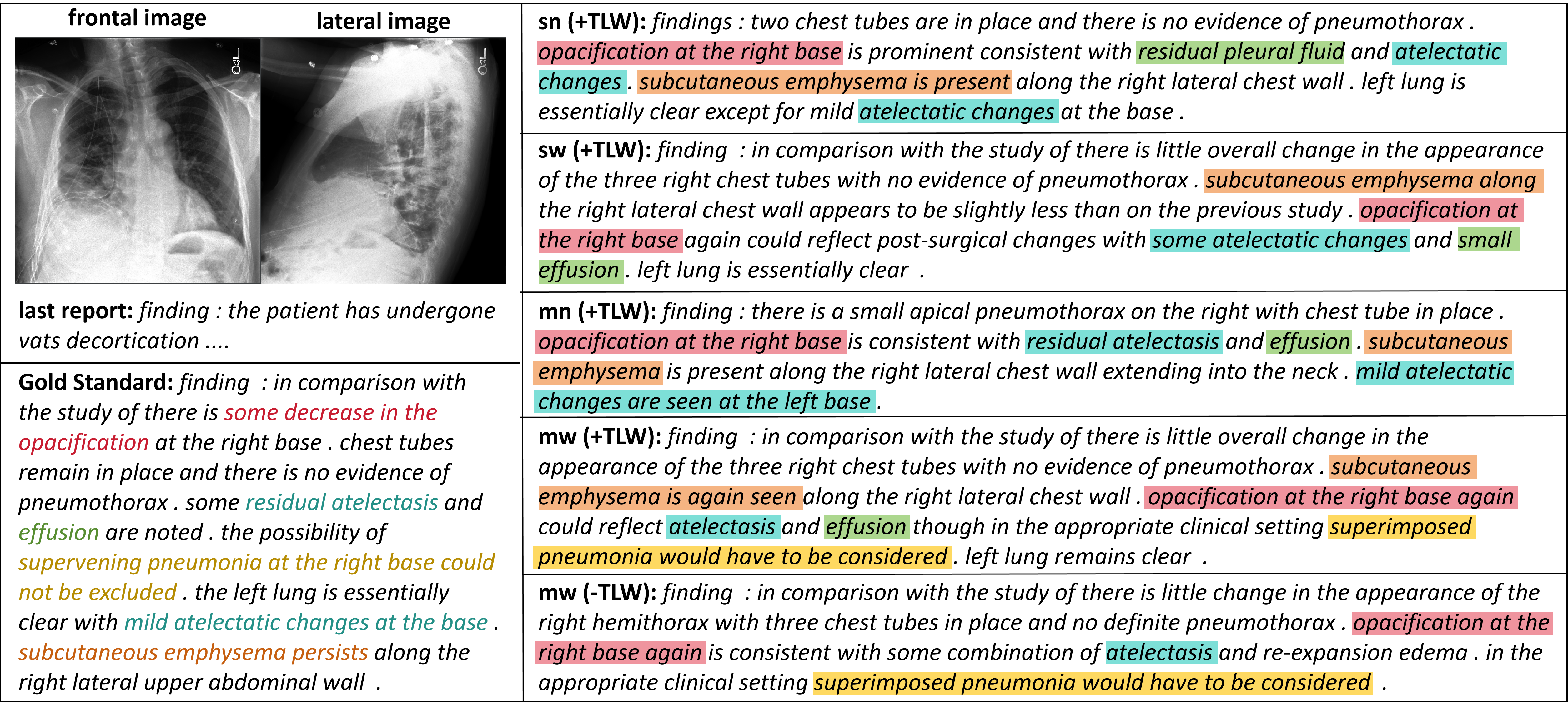} 
	\caption{ An illustration of a challenging case featuring five positive or uncertain diagnoses across four different settings, where the gold standard is also presented for reference. Diagnosis shared by the gold standard and model outputs are highlighted in the same color. LLM-RG4 identifies nearly all diagnoses, whereas the absence of TLW leads to the missing of two diagnoses.}
	\label{fig4}
\end{figure*}

\section{Experimental Setup}
\subsection{Datasets and Metrics}
We train LLM-RG4 on MIMIC-RG4 consisting of four input scenarios: single view no longitudinal, multi-view no longitudinal, single view with longitudinal, and multi-view with longitudinal (denoted as \textbf{\textit{sn}}, \textbf{\textit{sw}}, \textbf{\textit{mn}}, \textbf{\textit{mw}} below). We include the indication/history section as input when available.

Model performance is assessed across three dimensions: natural language generation (NLG), clinical efficacy (CE), and hallucination (\textbf{\textit{hall.}}). For NLG, we use BLEU (B@n) \cite{papineni2002bleu}, METEOR (MTR) \cite{banerjee2005meteor} and ROUGE-L (R-L) \cite{lin2004rouge}. 
For CE, we adopt CheXbert to extract category labels and calculate micro-averaged precision (P), recall (R), and F1-score (F1), following established settings~\cite{jin2024promptmrg}. For hallucinations, we emphasize input-agnostic hallucinations (\textbf{\textit{hall.}}) and employ DiscBERT to measure the proportion of generated reports containing input-agnostic information. A lower value of \textbf{\textit{hall.}} reflects a diminished occurrence of input-agnostic hallucination. Additionally, we use the Wilcoxon Signed-Rank Test \cite{woolson2005wilcoxon} to assess performance improvements over baselines. All reports are kept untruncated during testing.	

\subsection{Baselines}
Comparative experiments are performed on the traditional \textbf{\textit{sn}} task and the multi-task MIMIC-RG4. For traditional \textbf{\textit{sn}}, we adopt frontal images and focus solely on the findings section in MIMIC-CXR following \cite{chen2020generating}.
For MIMIC-RG4, both finding and impression are evaluated. We retrain the encoder-decoder model CXRMate \cite{nicolson2023longitudinal}, which handles four input scenarios and is trained with reinforcement learning. Additionally, we compare LLM-RG4 with RadFM \cite{wu2023towards}, a 14B radiology multimodal large language model capable of processing interleaved text and image inputs. 

\subsection{Implementation Details}
We adopt RAD-DINO \cite{perez2024rad} as the image encoder and BiomedVLP-CXRBERT \cite{boecking2022making} as the text encoder, with Vicuna 7B v1.5 \cite{chiang2023vicuna} as the text decoder. 
The number of learnable variable tokens in the perceiver is set to 128, threshold is set to 0.4 and \(\lambda\) is set to 1.75. Following LLAVA \cite{liu2024visual}, we employ a two-stage training strategy. Initially, we only train the ATF with \textbf{\textit{sn}} data to achieve modality alignment. Subsequently, we conduct instruction tuning on the MIMIC-RG4 dataset, training the ATF, and applying LoRA \cite{hu2021lora} for fine-tuning Vicuna.

\section{Results and Analysis}
\subsection{Overall Results}
Table 1 presents the experimental results on the conventional \textbf{\textit{sn}} task. Benefiting from the token-level loss weighting strategy, LLM-RG4 shows a 4.0\% and a 3.5\% absolute improvement respectively in F1 score over best classifier-assisted model Promptmrg and MAIRA-1 of the same 7B size using 1369 visual tokens. 
In clean NLG, where ground truth reports get 2.3\% \textbf{\textit{hall.}} score, LLM-RG4 has a 24.2\% relative improvement in BLEU-4 and a 39.2\% relative improvement in ROUGE-L, attributed to training on a cleaner dataset. Regarding hallucination, we find most open-source models suffer from hallucination problems. Even a multitasking model CheXagent also gets a 54.9\% \textbf{\textit{hall.}} score. However, LLM-RG4 achieves only a 1.5\% \textbf{\textit{hall.}} score, indicating that it exhibits minimal input-agnostic hallucinations.
Finally, we compute the original-NLG metrics with MIMIC-CXR reports for comparsion with powerful closed-source models.
Surprisingly, although the ground truth reports attain a 88.5\% \textbf{\textit{hall.}} score, our model remains competitive with SOTA models. Both BLEU-4 score and ROUGE-L score of LLM-RG4 are comparable to or slightly exceed the SOTA specialised large multimodal model MAIRA-1. The minimal \textbf{\textit{hall.}} suggests a substantial number of valid descriptions matching the ground truth. This result likely arises from our model being trained on a mixture of four kinds of clean reports, which enhance its learning of valid descriptions.

Under the MIMIC-RG4 setting, LLM-RG4 outperforms all existing models supporting MIMIC-RG4. Although cxrmate utilized reinforcement learning to enhance clinical accuracy, we still achieve an average absolute F1 improvement of 3.8\% across four tasks. LLM-RG4's superior performance on NLG metrics further demonstrates that the MLLM architecture is well-suited for multi-tasks featuring flexible language generaion. While RadFM can support interleaved text and image inputs, it performs unsatisfactorily and exhibits a significant input-agnostic hallucination issue on the \textbf{\textit{sn}} setting likely due to overfitting.

\subsection{Ablation Analysis}

\begin{table}[b!]
\centering
\footnotesize
\newcolumntype{C}{>{\centering\arraybackslash}X}%
\begin{tabularx}{\columnwidth}{CCCCCCCC}
    \toprule	
    \multirow{2}{*}{\textbf{stage}} & 
    \multirow{2}{*}{\textbf{ATF}} & \multirow{2}{*}{\textbf{TLW}} & \multirow{2}{*}{\textbf{N.}}& \multicolumn{2}{c}{\textbf{F1 Score}} &\multicolumn{2}{c}{\textbf{NLG}}\\ 
    \cmidrule(r){5-6} \cmidrule(r){7-8}
    & & & & F-14 & F-5 & B@1 & B@4\\
    \midrule
    1 & - & \xmark & 128 &0.519&0.563&0.388&0.126\\
    1 & - & \cmark & 128 &\textbf{0.580}&\textbf{0.619}&\textbf{0.400}&\textbf{0.129}\\
    \cmidrule(r){1-8}
    2 &\xmark& \xmark & 502 &0.551&0.582&0.468&0.201\\
    2 &\xmark& \cmark & 502 &0.577&0.603&0.469&0.197\\
    2 &\cmark& \xmark & \textbf{204} &0.556 &0.580&0.467&\textbf{0.205}\\
    2 &\cmark& \cmark & \textbf{204} &\textbf{0.585}&\textbf{0.610}&\textbf{0.472}&0.199\\
    \bottomrule
    \end{tabularx}
    \caption{Ablation study of each module on MIMIC-RG4. Stage2 results are the average scores across four settings. N. means the max input token numbers. }
    \label{table1}
\end{table}
We conduct ablation experiments of alternative model designs, as shown in Table 4.
The adaptive token fusion module helps reduce feature tokens by approximately 60\% when both multi-view and longitudinal information are available, offering similar or slightly better performance than original interleaved inputs. This supports our hypothesis that, efficient and high-fidelity encoding for MLLM can be achieved within specialized medical tasks. TLW benefits CE results on both ATF architecture and interleaved inputs architecture, suggesting that assigning higher loss weights to positive and uncertain descriptions during training can effectively enhance the model's focus on underlying lesions. This phenomenon is evident in both the first and second stages. 
Further analyses about the ATF module and TLW module are provided in the supplementary materials.

\subsection{Influence of Mixed Training}
To further investigate whether mixed training across the four settings yields improvement, we train each setting individually for the same number of epochs, shown in Table 5. 
We find that mixed training lead to varying degrees of improvement across different scenarios, particularly in challenging settings (\textbf{\textit{mn}}, \textbf{\textit{mw}}). We hypothesize that such mixed training can be regarded as a form of data augmentation, effectively increasing the diversity of the training data. It explicitly enables the model to learn the varying demands across different settings during training.
\begin{table}[h]
\centering
\footnotesize
\newcolumntype{C}{>{\centering\arraybackslash}X}%
\begin{tabularx}{\columnwidth}{CCCCCCCCC}
    \toprule	 
    \multirow{2}{*}{\textbf{Setting}} & \multirow{2}{*}{\textbf{T.S.}}
    & \multicolumn{3}{c}{\textbf{CE Metrics}} & \multicolumn{3}{c}{\textbf{NLG Metrics}}\\ 
    \cmidrule(r){3-5} \cmidrule(r){6-8}
    & &P & R & F1 & B@1 & B@4 & R-L\\
    \midrule
    \multirow{2}{*}{\textbf{\textit{sn}}}& \textbf{S} &\textbf{0.589}&0.609&0.599&0.456&0.186&0.384\\
    & \textbf{M} &0.588&\textbf{0.632}&\textbf{0.609}&\textbf{0.479}&\textbf{0.196}&\textbf{0.384}\\
    \cmidrule(r){1-8}
    \multirow{2}{*}{\textbf{\textit{sw}}}& \textbf{S} &0.596&0.588&0.592&0.427&0.173&0.376\\
    & \textbf{M} &\textbf{0.599}&\textbf{0.622}&\textbf{0.610}&\textbf{0.455}&\textbf{0.186}&\textbf{0.382}\\
    \cmidrule(r){1-8}
    \multirow{2}{*}{\textbf{\textit{mn}}}& \textbf{S} &0.526&0.531&0.528&0.464&0.202&0.384\\
    & \textbf{M} &\textbf{0.541}&\textbf{0.578}&\textbf{0.559}&\textbf{0.491}&\textbf{0.216}&\textbf{0.405}\\
    \cmidrule(r){1-8}
    \multirow{2}{*}{\textbf{\textit{mw}}}& \textbf{S} &0.519&0.544&0.531&0.446&0.179&0.384\\
    & \textbf{M} &\textbf{0.560}&\textbf{0.565}&\textbf{0.563}&\textbf{0.461}&\textbf{0.197}&\textbf{0.401}\\
    \bottomrule
    \end{tabularx}
    \caption{Influence of mixed training across four settings. T.S. represents the training strategy, \textbf{M} represents mixed training, \textbf{S} represents training on the specific setting.}
    \label{table1}
\end{table}
\subsection{Case study}
We present a qualitative example to illustrate LLM-RG4's flexibility of diverse inputs and investigate the impact of the TLW module on the model's capabilities, as shown in Figure 3. We select a challenging case involving a patient with confirmed or suspected diagnoses of five different diseases. In the context of multi-view X-ray inputs and longitudinal data, the reports generated by LLM-RG4 accurately cover all five diseases, demonstrating its clinical diagnostic accuracy. In the absence of the token-level loss weighting strategy (TLW), the generated reports lack two diagnostic information. 
Additionally, the comparative description appears only when longitudinal data is included (\textbf{\textit{sw}}, \textbf{\textit{mw}}), indicating consistency between model inputs and outputs. For the limitations of LLM-RG4, while it provides four types of diagnostic descriptions under scenarios such as \textbf{\textit{sn}}, \textbf{\textit{sw}}, and \textbf{\textit{mn}}, it does not mention the need to consider pneumonia. In the \textbf{\textit{mn}} scenario, it suspects the presence of a small pneumothorax. Future work should further constrain LLM-RG4 to ensure the consistency across different input scenarios. Such enhancement would be beneficial for clinical practice.

\section{Conclusion}
In this work, we introduce MIMIC-RG4, a novel paradigm for radiology report generation that adapts to varying input scenarios, aligning more closely with clinical report writing practices. We further propose ATF and TLW to enhance the flexibility and accuracy of large language models in handling diverse inputs, with a consistent emphasis on identifying pathological findings across different settings. Experiments conducted on two datasets illustrate the effectiveness of our method, highlighting that MLLM can achieve more compact information encoding for specific medical tasks. Additionally, the emphasis on key semantic tokens at the loss layer is crucial for enhancing clinical efficacy. We hope these efforts will provide new insights into radiology report generation and the application of large language models in biomedical domains.

\bibliography{aaai25}

\begin{thebibliography}{61}
\providecommand{\natexlab}[1]{#1}

\bibitem[{AI@Meta(2024)}]{llama3modelcard}
AI@Meta. 2024.
\newblock Llama 3 Model Card.

\bibitem[{Banerjee and Lavie(2005)}]{banerjee2005meteor}
Banerjee, S.; and Lavie, A. 2005.
\newblock METEOR: An automatic metric for MT evaluation with improved
  correlation with human judgments.
\newblock In \emph{Proceedings of the acl workshop on intrinsic and extrinsic
  evaluation measures for machine translation and/or summarization}, 65--72.

\bibitem[{Boecking et~al.(2022)Boecking, Usuyama, Bannur, Castro, Schwaighofer,
  Hyland, Wetscherek, Naumann, Nori, Alvarez-Valle et~al.}]{boecking2022making}
Boecking, B.; Usuyama, N.; Bannur, S.; Castro, D.~C.; Schwaighofer, A.; Hyland,
  S.; Wetscherek, M.; Naumann, T.; Nori, A.; Alvarez-Valle, J.; et~al. 2022.
\newblock Making the most of text semantics to improve biomedical
  vision--language processing.
\newblock In \emph{European conference on computer vision}, 1--21. Springer.

\bibitem[{Bu et~al.(2024)Bu, Li, Yang, and Dai}]{bu2024instance}
Bu, S.; Li, T.; Yang, Y.; and Dai, Z. 2024.
\newblock Instance-level Expert Knowledge and Aggregate Discriminative
  Attention for Radiology Report Generation.
\newblock In \emph{Proceedings of the IEEE/CVF Conference on Computer Vision
  and Pattern Recognition}, 14194--14204.

\bibitem[{Chen et~al.(2021)Chen, Shen, Song, and Wan}]{chen2021cross}
Chen, Z.; Shen, Y.; Song, Y.; and Wan, X. 2021.
\newblock Cross-modal Memory Networks for Radiology Report Generation.
\newblock In \emph{Proceedings of the 59th Annual Meeting of the Association
  for Computational Linguistics and the 11th International Joint Conference on
  Natural Language Processing (Volume 1: Long Papers)}, 5904--5914.

\bibitem[{Chen et~al.(2020)Chen, Song, Chang, and Wan}]{chen2020generating}
Chen, Z.; Song, Y.; Chang, T.-H.; and Wan, X. 2020.
\newblock Generating Radiology Reports via Memory-driven Transformer.
\newblock In \emph{Proceedings of the 2020 Conference on Empirical Methods in
  Natural Language Processing (EMNLP)}, 1439--1449.

\bibitem[{Chen et~al.(2024)Chen, Varma, Delbrouck, Paschali, Blankemeier,
  Van~Veen, Valanarasu, Youssef, Cohen, Reis et~al.}]{chen2024chexagent}
Chen, Z.; Varma, M.; Delbrouck, J.-B.; Paschali, M.; Blankemeier, L.; Van~Veen,
  D.; Valanarasu, J. M.~J.; Youssef, A.; Cohen, J.~P.; Reis, E.~P.; et~al.
  2024.
\newblock Chexagent: Towards a foundation model for chest x-ray interpretation.
\newblock \emph{arXiv preprint arXiv:2401.12208}.

\bibitem[{Chiang et~al.(2023)Chiang, Li, Lin, Sheng, Wu, Zhang, Zheng, Zhuang,
  Zhuang, Gonzalez et~al.}]{chiang2023vicuna}
Chiang, W.-L.; Li, Z.; Lin, Z.; Sheng, Y.; Wu, Z.; Zhang, H.; Zheng, L.;
  Zhuang, S.; Zhuang, Y.; Gonzalez, J.~E.; et~al. 2023.
\newblock Vicuna: An open-source chatbot impressing gpt-4 with 90\%* chatgpt
  quality.
\newblock \emph{See https://vicuna. lmsys. org (accessed 14 April 2023)}, 2(3):
  6.

\bibitem[{Dalla~Serra et~al.(2023)Dalla~Serra, Wang, Deligianni, Dalton, and
  O’Neil}]{dalla2023controllable}
Dalla~Serra, F.; Wang, C.; Deligianni, F.; Dalton, J.; and O’Neil, A. 2023.
\newblock Controllable Chest X-Ray Report Generation from Longitudinal
  Representations.
\newblock In \emph{Findings of the Association for Computational Linguistics:
  EMNLP 2023}, 4891--4904.

\bibitem[{Deria et~al.(2024)Deria, Kumar, Chakraborty, Mahapatra, and
  Roy}]{deria2024inverge}
Deria, A.; Kumar, K.; Chakraborty, S.; Mahapatra, D.; and Roy, S. 2024.
\newblock InVERGe: Intelligent Visual Encoder for Bridging Modalities in Report
  Generation.
\newblock In \emph{Proceedings of the IEEE/CVF Conference on Computer Vision
  and Pattern Recognition}, 2028--2038.

\bibitem[{Devlin(2018)}]{devlin2018bert}
Devlin, J. 2018.
\newblock Bert: Pre-training of deep bidirectional transformers for language
  understanding.
\newblock \emph{arXiv preprint arXiv:1810.04805}.

\bibitem[{Guo et~al.(2023{\natexlab{a}})Guo, Tang, Zhang, Wang, Wang, Zhao, and
  Li}]{guo2023viewrefer}
Guo, Z.; Tang, Y.; Zhang, R.; Wang, D.; Wang, Z.; Zhao, B.; and Li, X.
  2023{\natexlab{a}}.
\newblock Viewrefer: Grasp the multi-view knowledge for 3d visual grounding.
\newblock In \emph{Proceedings of the IEEE/CVF International Conference on
  Computer Vision}, 15372--15383.

\bibitem[{Guo et~al.(2023{\natexlab{b}})Guo, Zhang, Zhu, Tang, Ma, Han, Chen,
  Gao, Li, Li et~al.}]{guo2023point}
Guo, Z.; Zhang, R.; Zhu, X.; Tang, Y.; Ma, X.; Han, J.; Chen, K.; Gao, P.; Li,
  X.; Li, H.; et~al. 2023{\natexlab{b}}.
\newblock Point-bind \& point-llm: Aligning point cloud with multi-modality for
  3d understanding, generation, and instruction following.
\newblock \emph{arXiv preprint arXiv:2309.00615}.

\bibitem[{Hartung et~al.(2020)Hartung, Bickle, Gaillard, and
  Kanne}]{hartung2020create}
Hartung, M.~P.; Bickle, I.~C.; Gaillard, F.; and Kanne, J.~P. 2020.
\newblock How to create a great radiology report.
\newblock \emph{Radiographics}, 40(6): 1658--1670.

\bibitem[{Hu et~al.(2021)Hu, Shen, Wallis, Allen-Zhu, Li, Wang, Wang, and
  Chen}]{hu2021lora}
Hu, E.~J.; Shen, Y.; Wallis, P.; Allen-Zhu, Z.; Li, Y.; Wang, S.; Wang, L.; and
  Chen, W. 2021.
\newblock Lora: Low-rank adaptation of large language models.
\newblock \emph{arXiv preprint arXiv:2106.09685}.

\bibitem[{Huang, Zhang, and Zhang(2023)}]{huang2023kiut}
Huang, Z.; Zhang, X.; and Zhang, S. 2023.
\newblock Kiut: Knowledge-injected u-transformer for radiology report
  generation.
\newblock In \emph{Proceedings of the IEEE/CVF Conference on Computer Vision
  and Pattern Recognition}, 19809--19818.

\bibitem[{Hyland et~al.(2023)Hyland, Bannur, Bouzid, Castro, Ranjit,
  Schwaighofer, P{\'e}rez-Garc{\'\i}a, Salvatelli, Srivastav, Thieme
  et~al.}]{hyland2023maira}
Hyland, S.~L.; Bannur, S.; Bouzid, K.; Castro, D.~C.; Ranjit, M.; Schwaighofer,
  A.; P{\'e}rez-Garc{\'\i}a, F.; Salvatelli, V.; Srivastav, S.; Thieme, A.;
  et~al. 2023.
\newblock Maira-1: A specialised large multimodal model for radiology report
  generation.
\newblock \emph{arXiv preprint arXiv:2311.13668}.

\bibitem[{Jaegle et~al.(2021)Jaegle, Gimeno, Brock, Vinyals, Zisserman, and
  Carreira}]{jaegle2021perceiver}
Jaegle, A.; Gimeno, F.; Brock, A.; Vinyals, O.; Zisserman, A.; and Carreira, J.
  2021.
\newblock Perceiver: General perception with iterative attention.
\newblock In \emph{International conference on machine learning}, 4651--4664.
  PMLR.

\bibitem[{Jin et~al.(2024)Jin, Che, Lin, and Chen}]{jin2024promptmrg}
Jin, H.; Che, H.; Lin, Y.; and Chen, H. 2024.
\newblock Promptmrg: Diagnosis-driven prompts for medical report generation.
\newblock In \emph{Proceedings of the AAAI Conference on Artificial
  Intelligence}, volume~38, 2607--2615.

\bibitem[{Johnson et~al.(2019)Johnson, Pollard, Berkowitz, Greenbaum, Lungren,
  Deng, Mark, and Horng}]{johnson2019mimic}
Johnson, A.~E.; Pollard, T.~J.; Berkowitz, S.~J.; Greenbaum, N.~R.; Lungren,
  M.~P.; Deng, C.-y.; Mark, R.~G.; and Horng, S. 2019.
\newblock MIMIC-CXR, a de-identified publicly available database of chest
  radiographs with free-text reports.
\newblock \emph{Scientific data}, 6(1): 317.

\bibitem[{Lee et~al.(2023)Lee, Kim, Kim, Kim, Kim, Sunwoo, and
  Choi}]{lee2023unified}
Lee, H.; Kim, W.; Kim, J.-H.; Kim, T.; Kim, J.; Sunwoo, L.; and Choi, E. 2023.
\newblock Unified chest x-ray and radiology report generation model with
  multi-view chest x-rays.
\newblock \emph{arXiv preprint arXiv:2302.12172}, 3(7): 8.

\bibitem[{Li et~al.(2023{\natexlab{a}})Li, Zhang, Chen, Wang, Pu, Yang, Li, and
  Liu}]{li2023mimic}
Li, B.; Zhang, Y.; Chen, L.; Wang, J.; Pu, F.; Yang, J.; Li, C.; and Liu, Z.
  2023{\natexlab{a}}.
\newblock Mimic-it: Multi-modal in-context instruction tuning.
\newblock \emph{arXiv preprint arXiv:2306.05425}.

\bibitem[{Li et~al.(2023{\natexlab{b}})Li, Li, Savarese, and Hoi}]{li2023blip}
Li, J.; Li, D.; Savarese, S.; and Hoi, S. 2023{\natexlab{b}}.
\newblock BLIP-2: bootstrapping language-image pre-training with frozen image
  encoders and large language models.
\newblock In \emph{Proceedings of the 40th International Conference on Machine
  Learning}, 19730--19742.

\bibitem[{Li et~al.(2023{\natexlab{c}})Li, Lin, Chen, Lin, Liang, and
  Chang}]{li2023dynamic}
Li, M.; Lin, B.; Chen, Z.; Lin, H.; Liang, X.; and Chang, X.
  2023{\natexlab{c}}.
\newblock Dynamic graph enhanced contrastive learning for chest x-ray report
  generation.
\newblock In \emph{Proceedings of the IEEE/CVF Conference on Computer Vision
  and Pattern Recognition}, 3334--3343.

\bibitem[{Li et~al.(2018)Li, Liang, Hu, and Xing}]{li2018hybrid}
Li, Y.; Liang, X.; Hu, Z.; and Xing, E.~P. 2018.
\newblock Hybrid retrieval-generation reinforced agent for medical image report
  generation.
\newblock \emph{Advances in neural information processing systems}, 31.

\bibitem[{Lin(2004)}]{lin2004rouge}
Lin, C.-Y. 2004.
\newblock Rouge: A package for automatic evaluation of summaries.
\newblock In \emph{Text summarization branches out}, 74--81.

\bibitem[{Lin et~al.(2024)Lin, Gou, Gong, Liu, Shen, Xu, Lin, Yang, Jiao, Duan
  et~al.}]{lin2024rho}
Lin, Z.; Gou, Z.; Gong, Y.; Liu, X.; Shen, Y.; Xu, R.; Lin, C.; Yang, Y.; Jiao,
  J.; Duan, N.; et~al. 2024.
\newblock Rho-1: Not all tokens are what you need.
\newblock \emph{arXiv preprint arXiv:2404.07965}.

\bibitem[{Liu et~al.(2024{\natexlab{a}})Liu, Tian, Chen, Song, and
  Zhang}]{liu2024bootstrapping}
Liu, C.; Tian, Y.; Chen, W.; Song, Y.; and Zhang, Y. 2024{\natexlab{a}}.
\newblock Bootstrapping Large Language Models for Radiology Report Generation.
\newblock In \emph{Proceedings of the AAAI Conference on Artificial
  Intelligence}, volume~38, 18635--18643.

\bibitem[{Liu et~al.(2024{\natexlab{b}})Liu, Li, Wu, and Lee}]{liu2024visual}
Liu, H.; Li, C.; Wu, Q.; and Lee, Y.~J. 2024{\natexlab{b}}.
\newblock Visual instruction tuning.
\newblock \emph{Advances in neural information processing systems}, 36.

\bibitem[{Miura et~al.(2021)Miura, Zhang, Tsai, Langlotz, and
  Jurafsky}]{miura2021improving}
Miura, Y.; Zhang, Y.; Tsai, E.; Langlotz, C.; and Jurafsky, D. 2021.
\newblock Improving Factual Completeness and Consistency of Image-to-Text
  Radiology Report Generation.
\newblock In \emph{Proceedings of the 2021 Conference of the North American
  Chapter of the Association for Computational Linguistics: Human Language
  Technologies}, 5288--5304.

\bibitem[{Nguyen et~al.(2023)Nguyen, Chen, He, and Tan}]{nguyen2023pragmatic}
Nguyen, D.; Chen, C.; He, H.; and Tan, C. 2023.
\newblock Pragmatic radiology report generation.
\newblock In \emph{Machine Learning for Health (ML4H)}, 385--402. PMLR.

\bibitem[{Nguyen et~al.(2021)Nguyen, Nie, Badamdorj, Liu, Zhu, Truong, and
  Cheng}]{nguyen2021automated}
Nguyen, H.~T.; Nie, D.; Badamdorj, T.; Liu, Y.; Zhu, Y.; Truong, J.; and Cheng,
  L. 2021.
\newblock Automated generation of accurate$\backslash$\& fluent medical x-ray
  reports.
\newblock \emph{arXiv preprint arXiv:2108.12126}.

\bibitem[{Nicolson, Dowling, and
  Koopman(2023{\natexlab{a}})}]{nicolson2023improving}
Nicolson, A.; Dowling, J.; and Koopman, B. 2023{\natexlab{a}}.
\newblock Improving chest X-ray report generation by leveraging warm starting.
\newblock \emph{Artificial intelligence in medicine}, 144: 102633.

\bibitem[{Nicolson, Dowling, and
  Koopman(2023{\natexlab{b}})}]{nicolson2023longitudinal}
Nicolson, A.; Dowling, J.; and Koopman, B. 2023{\natexlab{b}}.
\newblock Longitudinal Data and a Semantic Similarity Reward for Chest X-Ray
  Report Generation.
\newblock \emph{arXiv preprint arXiv:2307.09758}.

\bibitem[{Papineni et~al.(2002)Papineni, Roukos, Ward, and
  Zhu}]{papineni2002bleu}
Papineni, K.; Roukos, S.; Ward, T.; and Zhu, W.-J. 2002.
\newblock Bleu: a method for automatic evaluation of machine translation.
\newblock In \emph{Proceedings of the 40th annual meeting of the Association
  for Computational Linguistics}, 311--318.

\bibitem[{P{\'e}rez-Garc{\'\i}a et~al.(2024)P{\'e}rez-Garc{\'\i}a, Sharma,
  Bond-Taylor, Bouzid, Salvatelli, Ilse, Bannur, Castro, Schwaighofer, Lungren
  et~al.}]{perez2024rad}
P{\'e}rez-Garc{\'\i}a, F.; Sharma, H.; Bond-Taylor, S.; Bouzid, K.; Salvatelli,
  V.; Ilse, M.; Bannur, S.; Castro, D.~C.; Schwaighofer, A.; Lungren, M.~P.;
  et~al. 2024.
\newblock RAD-DINO: Exploring Scalable Medical Image Encoders Beyond Text
  Supervision.
\newblock \emph{arXiv preprint arXiv:2401.10815}.

\bibitem[{Ramesh, Chi, and Rajpurkar(2022)}]{ramesh2022improving}
Ramesh, V.; Chi, N.~A.; and Rajpurkar, P. 2022.
\newblock Improving radiology report generation systems by removing
  hallucinated references to non-existent priors.
\newblock In \emph{Machine Learning for Health}, 456--473. PMLR.

\bibitem[{Sanjeev et~al.(2024)Sanjeev, Maani, Abzhanov, Papineni, Almakky,
  Papie{\.z}, and Yaqub}]{sanjeev2024tibix}
Sanjeev, S.; Maani, F.~A.; Abzhanov, A.; Papineni, V.~R.; Almakky, I.;
  Papie{\.z}, B.~W.; and Yaqub, M. 2024.
\newblock TiBiX: Leveraging Temporal Information for Bidirectional X-ray and
  Report Generation.
\newblock \emph{arXiv preprint arXiv:2403.13343}.

\bibitem[{Serra et~al.(2023)Serra, Wang, Deligianni, Dalton, and
  O'Neil}]{serra2023controllable}
Serra, F.~D.; Wang, C.; Deligianni, F.; Dalton, J.; and O'Neil, A.~Q. 2023.
\newblock Controllable chest x-ray report generation from longitudinal
  representations.
\newblock \emph{arXiv preprint arXiv:2310.05881}.

\bibitem[{Smit et~al.(2020)Smit, Jain, Rajpurkar, Pareek, Ng, and
  Lungren}]{smit2020combining}
Smit, A.; Jain, S.; Rajpurkar, P.; Pareek, A.; Ng, A.~Y.; and Lungren, M. 2020.
\newblock Combining Automatic Labelers and Expert Annotations for Accurate
  Radiology Report Labeling Using BERT.
\newblock In \emph{Proceedings of the 2020 Conference on Empirical Methods in
  Natural Language Processing (EMNLP)}, 1500--1519.

\bibitem[{Sun et~al.(2024)Sun, Khor, Wang, Wang, Zhao, Zhang, Ma, Zheng, and
  Liao}]{sun2024continually}
Sun, Y.; Khor, H.~G.; Wang, Y.; Wang, Z.; Zhao, H.; Zhang, Y.; Ma, L.; Zheng,
  Z.; and Liao, H. 2024.
\newblock Continually Tuning a Large Language Model for Multi-domain Radiology
  Report Generation.
\newblock In \emph{International Conference on Medical Image Computing and
  Computer-Assisted Intervention}, 177--187. Springer.

\bibitem[{Sundararajan, Taly, and Yan(2017)}]{sundararajan2017axiomatic}
Sundararajan, M.; Taly, A.; and Yan, Q. 2017.
\newblock Axiomatic attribution for deep networks.
\newblock In \emph{International conference on machine learning}, 3319--3328.
  PMLR.

\bibitem[{Tang et~al.(2024)Tang, Zhang, Guo, Ma, Zhao, Wang, Wang, and
  Li}]{tang2024point}
Tang, Y.; Zhang, R.; Guo, Z.; Ma, X.; Zhao, B.; Wang, Z.; Wang, D.; and Li, X.
  2024.
\newblock Point-PEFT: Parameter-efficient fine-tuning for 3D pre-trained
  models.
\newblock In \emph{Proceedings of the AAAI Conference on Artificial
  Intelligence}, volume~38, 5171--5179.

\bibitem[{Tanida et~al.(2023)Tanida, M{\"u}ller, Kaissis, and
  Rueckert}]{tanida2023interactive}
Tanida, T.; M{\"u}ller, P.; Kaissis, G.; and Rueckert, D. 2023.
\newblock Interactive and explainable region-guided radiology report
  generation.
\newblock In \emph{Proceedings of the IEEE/CVF Conference on Computer Vision
  and Pattern Recognition}, 7433--7442.

\bibitem[{Thawakar et~al.(2024)Thawakar, Shaker, Mullappilly, Cholakkal, Anwer,
  Khan, Laaksonen, and Khan}]{thawakar2024xraygpt}
Thawakar, O.~C.; Shaker, A.~M.; Mullappilly, S.~S.; Cholakkal, H.; Anwer,
  R.~M.; Khan, S.; Laaksonen, J.; and Khan, F. 2024.
\newblock XrayGPT: Chest Radiographs Summarization using Large Medical
  Vision-Language Models.
\newblock In \emph{Proceedings of the 23rd Workshop on Biomedical Natural
  Language Processing}, 440--448.

\bibitem[{Tu et~al.(2024)Tu, Azizi, Driess, Schaekermann, Amin, Chang, Carroll,
  Lau, Tanno, Ktena et~al.}]{tu2024towards}
Tu, T.; Azizi, S.; Driess, D.; Schaekermann, M.; Amin, M.; Chang, P.-C.;
  Carroll, A.; Lau, C.; Tanno, R.; Ktena, I.; et~al. 2024.
\newblock Towards generalist biomedical AI.
\newblock \emph{NEJM AI}, 1(3): AIoa2300138.

\bibitem[{Wang, Bhalerao, and He(2022)}]{wang2022cross}
Wang, J.; Bhalerao, A.; and He, Y. 2022.
\newblock Cross-modal prototype driven network for radiology report generation.
\newblock In \emph{European Conference on Computer Vision}, 563--579. Springer.

\bibitem[{Wang et~al.(2023{\natexlab{a}})Wang, Zhao, Ouyang, Wang, and
  Shen}]{wang2023chatcad}
Wang, S.; Zhao, Z.; Ouyang, X.; Wang, Q.; and Shen, D. 2023{\natexlab{a}}.
\newblock ChatCAD: interactive computer-aided diagnosis on medical image using
  large language models. arXiv. doi: 10.48550.
\newblock \emph{arXiv preprint arXiv.2302.07257}.

\bibitem[{Wang et~al.(2022{\natexlab{a}})Wang, Han, Wang, Li, and
  Zhou}]{wang2022automated}
Wang, Z.; Han, H.; Wang, L.; Li, X.; and Zhou, L. 2022{\natexlab{a}}.
\newblock Automated radiographic report generation purely on transformer: A
  multicriteria supervised approach.
\newblock \emph{IEEE Transactions on Medical Imaging}, 41(10): 2803--2813.

\bibitem[{Wang et~al.(2023{\natexlab{b}})Wang, Liu, Wang, and
  Zhou}]{wang2023metransformer}
Wang, Z.; Liu, L.; Wang, L.; and Zhou, L. 2023{\natexlab{b}}.
\newblock Metransformer: Radiology report generation by transformer with
  multiple learnable expert tokens.
\newblock In \emph{Proceedings of the IEEE/CVF Conference on Computer Vision
  and Pattern Recognition}, 11558--11567.

\bibitem[{Wang et~al.(2023{\natexlab{c}})Wang, Liu, Wang, and
  Zhou}]{wang2023r2gengpt}
Wang, Z.; Liu, L.; Wang, L.; and Zhou, L. 2023{\natexlab{c}}.
\newblock R2gengpt: Radiology report generation with frozen llms.
\newblock \emph{Meta-Radiology}, 1(3): 100033.

\bibitem[{Wang et~al.(2022{\natexlab{b}})Wang, Tang, Wang, Li, and
  Zhou}]{wang2022medical}
Wang, Z.; Tang, M.; Wang, L.; Li, X.; and Zhou, L. 2022{\natexlab{b}}.
\newblock A medical semantic-assisted transformer for radiographic report
  generation.
\newblock In \emph{International Conference on Medical Image Computing and
  Computer-Assisted Intervention}, 655--664. Springer.

\bibitem[{Woolson(2005)}]{woolson2005wilcoxon}
Woolson, R.~F. 2005.
\newblock Wilcoxon signed-rank test.
\newblock \emph{Encyclopedia of Biostatistics}, 8.

\bibitem[{Wu et~al.(2023)Wu, Zhang, Zhang, Wang, and Xie}]{wu2023towards}
Wu, C.; Zhang, X.; Zhang, Y.; Wang, Y.; and Xie, W. 2023.
\newblock Towards generalist foundation model for radiology.
\newblock \emph{arXiv preprint arXiv:2308.02463}.

\bibitem[{Wu, Huang, and Huang(2023)}]{wu2023token}
Wu, Y.; Huang, I.-C.; and Huang, X. 2023.
\newblock Token Imbalance Adaptation for Radiology Report Generation.
\newblock In \emph{Conference on Health, Inference, and Learning}, 72--85.
  PMLR.

\bibitem[{Xiao et~al.(2024)Xiao, Wu, Wang, Li, Zhou, and Guo}]{xiao2024seeing}
Xiao, X.; Wu, B.; Wang, J.; Li, C.; Zhou, X.; and Guo, H. 2024.
\newblock Seeing the Image: Prioritizing Visual Correlation by Contrastive
  Alignment.
\newblock \emph{arXiv preprint arXiv:2405.17871}.

\bibitem[{Yan et~al.(2021)Yan, He, Lu, Du, Chang, Gentili, McAuley, and
  Hsu}]{yan2021weakly}
Yan, A.; He, Z.; Lu, X.; Du, J.; Chang, E.; Gentili, A.; McAuley, J.; and Hsu,
  C.-N. 2021.
\newblock Weakly supervised contrastive learning for chest x-ray report
  generation.
\newblock \emph{arXiv preprint arXiv:2109.12242}.

\bibitem[{Yan et~al.(2024)Yan, Cheung, Tsang, Chiu, Tong, Cheung, and
  See}]{yan2024ahive}
Yan, S.; Cheung, W.~K.; Tsang, I.~W.; Chiu, K.; Tong, T.~M.; Cheung, K.~C.; and
  See, S. 2024.
\newblock AHIVE: Anatomy-aware Hierarchical Vision Encoding for Interactive
  Radiology Report Retrieval.
\newblock In \emph{Proceedings of the IEEE/CVF Conference on Computer Vision
  and Pattern Recognition}, 14324--14333.

\bibitem[{Yuan et~al.(2019)Yuan, Liao, Luo, and Luo}]{yuan2019automatic}
Yuan, J.; Liao, H.; Luo, R.; and Luo, J. 2019.
\newblock Automatic radiology report generation based on multi-view image
  fusion and medical concept enrichment.
\newblock In \emph{Medical Image Computing and Computer Assisted
  Intervention--MICCAI 2019: 22nd International Conference, Shenzhen, China,
  October 13--17, 2019, Proceedings, Part VI 22}, 721--729. Springer.

\bibitem[{Zhang et~al.(2020)Zhang, Wang, Xu, Yu, Yuille, and
  Xu}]{zhang2020radiology}
Zhang, Y.; Wang, X.; Xu, Z.; Yu, Q.; Yuille, A.; and Xu, D. 2020.
\newblock When radiology report generation meets knowledge graph.
\newblock In \emph{Proceedings of the AAAI conference on artificial
  intelligence}, volume~34, 12910--12917.

\bibitem[{Zhao et~al.(2024)Zhao, Wang, Gu, Zhu, Mei, Zhuang, Cui, Wang, and
  Shen}]{zhao2024chatcad+}
Zhao, Z.; Wang, S.; Gu, J.; Zhu, Y.; Mei, L.; Zhuang, Z.; Cui, Z.; Wang, Q.;
  and Shen, D. 2024.
\newblock Chatcad+: Towards a universal and reliable interactive cad using
  llms.
\newblock \emph{IEEE Transactions on Medical Imaging}.

\end{thebibliography}
\clearpage
\ifodd\value{page}
\hbox{} 
\fi

\appendix
\section{Appendix}
\subsection{Related Works}
\subsubsection{Different Paradigm in CXR}
CXR report generation tasks typically produce the findings section from a single image, treating lateral views as independent samples~\cite{chen2020generating,chen2021cross}. These tasks often employ encoder-decoder architectures that follow the image captioning paradigm and improve generated reports by optimizing network structures~\cite{wang2022cross,wang2023metransformer,huang2023kiut,yan2021weakly} or incorporating external prior information~\cite{zhang2020radiology,li2023dynamic}. \citet{nguyen2021automated} investigated the impact of different view scans and histories on report generation. \citet{serra2023controllable} modeled longitudinal historical information, introducing a framework to align, concatenate, and fuse current and prior scans into a joint representation. \citet{sanjeev2024tibix} and \citet{lee2023unified} attempted to merge report and image generation within a unified framework. Our paradigm focuses on clinical practice and emphasizes the flexibility and realism of report generation. It requires a unified model that generate various, factual reports across real-world scenarios (w or w/o multi-view, w or w/o longitudinal data) for the same patient.
\subsubsection{Large Language Model for Report Generation}
The advent of large-scale language models has significantly advanced the field of visual language~\cite{liu2024visual,li2023blip,guo2023viewrefer,tang2024point}. Numerous studies have leveraged these models for radiology report generation. Compared to training models from scratch on small datasets, using a large language model as a text decoder effectively utilizes its pretrained capabilities on large-scale datasets. \citet{wang2023chatcad,zhao2024chatcad+} translated results from computer-aided diagnosis networks into linguistic form, enabling integration of outputs from various networks. \citet{wang2023r2gengpt} and \citet{thawakar2024xraygpt} adopted visual tuning instructions, freezing large language models, and fine-tuning image encoders or connectors. \citet{liu2024bootstrapping} proposed bootstrapping LLMs for RRG with in-domain instance induction and a coarse-to-fine decoding process. \citet{deria2024inverge} presented a Cross-Modal Query Fusion Layer to enhance alignment between vision and language. \citet{hyland2023maira} explored key components affecting model performance, such as domain-specific image encoders, data augmentation, and the indications section. \citet{sun2024continually} proposed a continual learning approach tailored for report generation to address the challenge of catastrophic forgetting. 
Current large model-based report generation generally follows the format of visual instruction tuning. However, this format increases the token number with more inputs, leading to a greater computational burden. Our adaptive token fusion addresses this by using the visual token of the current main view as a query, interacting with other information to obtain fusion feature. This approach ensures that the different information interacts with each other and does not generate additional tokens. Reports can also be generated directly using the main view visual token when no other information is available.
\subsubsection{Token-level Loss Weighting Strategy}
Each term in a radiology report carries a distinct level of significance and should not be treated uniformly. \citet{wang2022automated} utilized the TF-IDF metric to measure the frequency of occurrence of each word and assigned different weights. \citet{wu2023token} penalized frequently generated tokens while using reinforcement learning to constantly update the range of frequent tokens. Beyond word frequency, recent NLP research has highlighted the importance of directly considering token semantics. \citet{lin2024rho} suggested that not all tokens are equally useful and proposed selective language modeling to focus on tokens consistent with the desired distribution, thus improving model performance on noisy datasets. \citet{xiao2024seeing} determined the visual relevance of each text token by comparing them with image inputs and assigned different loss weights accordingly. 
Considering our objective to prioritize positive or uncertain descriptions across various input scenarios, it is more significant to concentrate on the token semantic level rather than merely focusing on word frequency.
We propose leveraging a combination of CheXbert and Integral Gradients to automatically identify key diagnostic tokens within each report and assign them higher weights, without incurring any additional computational burden during training.

\begin{table*}[t]
\centering
\footnotesize
\newcolumntype{C}{>{\centering\arraybackslash}X}%
\begin{tabularx}{\linewidth}{lCCCCCCCCCCCC}
\toprule	 
\multirow{2}{*}{\textbf{Method}} & \multicolumn{3}{c}{\textbf{Prior Comparison}} & \multicolumn{3}{c}{\textbf{Prior Procedure }} &\multicolumn{3}{c}{\textbf{View }} & \multicolumn{3}{c}{\textbf{Communication}}\\ 
\cmidrule(r){2-4} \cmidrule(r){5-7} \cmidrule(r){8-10} \cmidrule(r){11-13} 
&P & R & F1 & P & R & F1 & P & R & F1 &P & R & F1\\
\midrule
Keywords&90.5&\textbf{99.3}&94.7&88.9&13.8&23.9&75.0&82.5&78.6&55.8&57.1&56.5 \\
Llama3&97.4&96.7&97.0&92.5&\textbf{84.5}&\textbf{88.1}&100.0&100.0&100.0&\textbf{97.5}&97.5&97.5 \\
DiscBERT&\textbf{97.4}&97.4&\textbf{97.4}&\textbf{93.9}&79.3&86.0&\textbf{100.0}&\textbf{100.0}&\textbf{100.0}&95.9&\textbf{100.0}&\textbf{97.9} \\
\bottomrule
\end{tabularx}
\caption{Comparison of various methods for quantitatively evaluating information categories in reports.}
\label{tab:Comparison_evaluating}
\end{table*}

\subsection{Details of MIMIC-RG4}
\subsubsection{Prompts for Llama3} We meticulously design prompts to instruct the Llama3 in judging reports, focusing primarily on the following four categories of information: prior comparison, prior procedure, multi-view description and communication description, as shown in Tabel 15. In prior comparison judgement rule, we incorporate the keywords used by \citet{nguyen2023pragmatic} as a reference for Llama3 and emphasize caution with implicit comparisons. We observe that the model often overlooks the term ``stable", so we highlight that it is derived from comparisons with previous reports. The prior procedure is the most challenging to assess, primarily due to the complexity of its context. Therefore, we clearly define three scenarios outlined in Table 15, including references to non-X-ray examinations, mentions of specific procedures, and the removal of specific instruments. For multi-view description and communication description, we alert Llama3 that suggestions don't fall into either category, which is frequently misclassified otherwise.

Comprehensive prompts for instructing LlaMA3 in reconstructing reports are provided in Table 16. We define reconstruction rules for common expressions, and the model is required to report any instances that fall outside these guidelines. 
The reconstruction rules primarily involve deleting or rewriting descriptions not inherent to the input, without introducing additional content. The sole exception is for multi-view descriptions. If multi-view descriptions are present, we add the sentence ``The lateral view is recommended to assist in diagnosis" after reconstruction.
The reconstruction instruction is combined with judgement instruction and judgement result to utilize LLM's chain-of-thought capability. In complex cases, a single modification may miss some expressions. Therefore, iterative modification rules are employed to facilitate multiple rounds of refinement.

\begin{table}[b!]
\centering
\footnotesize
\newcolumntype{C}{>{\centering\arraybackslash}X}%
\begin{tabularx}{\columnwidth}{lCCCCCCCC}
    \toprule	 
    Data & PC &PP & View &Comm& B@2 & B@4 &R\text{-}L& MTR\\
    \midrule
    \textit{Ori.} & 77.0 & 29.0 & 25.0 &20.0 & - & - & - & - \\
    \textit{Re.} &\textbf{2.5}&\textbf{2.5}&\textbf{5.5}&\textbf{0.0}&0.755&0.688&0.692&0.480\\
    \bottomrule
    \end{tabularx}
    \caption{Reconstruction evaluation, including percentage of uninferable information and NLG metrics. PC: Prior Comparison; PP: Prior Procedure; Comm: Communication; \textit{Ori.}: Original report; \textit{Re.}: Reconstructed report.}
    \label{table1}
\end{table}

\subsubsection{Implementation Details}
Doctors manually label and reconstruct 200 reports as a test set, where both the finding section and the impression section are included. The 200 reconstruction samples are processed to meet the single view no longitudinal criteria, employing all four types of judgement and reconstruction prompts. As such cases encompass the other three processing types, evaluation can be conducted based on this subset alone.
We choose Llama3-70B-Q4 model and deploy it on 4 RTX 3090 GPUs using Ollama. 
To achieve a balanced distribution of training samples for DiscBERT, we sample 11250 instances from each of the four configurations from MIMIC-CXR, totaling 45000 samples for the training set. We also select 1250 instances from each configuration, resulting in 5000 samples for the validation set. The manually annotated 200-sample test set is excluded.
DiscBERT adopts a modification of the BERT-base architecture \citet{devlin2018bert} with 4 linear heads, which correspond to four categories. We utilize cross-entropy loss and Adam optimization with a learning rate of $2 \times 10^{-5}$. During training, we periodically evaluate DiscBERT on the validation set and save the checkpoint with the highest performance averaged over all 4 categories. DiscBERT is trained using a RTX 3090 GPU with a batch size of 16.

\subsubsection{Pipeline Evaluation}
For DiscBERT evaluation, we compute precision (P), recall (R) and F1 score across four categories of information, as shown in Table 6. Keywords defined by \citet{nguyen2023pragmatic} are also used for comparison. 
We find that directly using keywords for classification poses challenges in balancing precision and recall due to linguistic ambiguity. A broader keyword selection enhances recall but reduces precision, as in the case of ``prior comparison", while a narrower selection improves precision but lowers recall, as with ``prior procedure". In contrast, Llama3 achieves a more balanced performance across metrics and generally surpasses keywords-based methods. DiscBERT, trained on Llama3-generated pseudo-labels, match Llama3's performance while drastically improving inference speed, processing 200 cases in nearly 1 second compared to Llama3's 2 minutes.

For reconstruction evaluation, we compute the percentage of reports containing different category description before and after the pipeline under the setting of \textbf{\textit{sn}}, using DiscBERT. We additionally 
calculate NLG metrics (B@2,B@4, R-L, MTR) using the LLM-reconstructed reports and manually-reconstructed reports to evaluate the similarity at the token level, as shown in Table 7. The pipeline substantially reduce uninferable information in the reports, and the carefully designed reconstruction rules also ensure a high token level similarity to manually-reconstructed reports. A qualitative result is illustrated in Figure 7.

\subsubsection{Complete Dataset Statistics}
The dataset statistics for \textbf{\textit{sw}}, \textbf{\textit{mn}}, and \textbf{\textit{mw}} are provided in Tables 8, 9, and 10. We observe two issues when directly filtering data from MIMIC-CXR based on four categories. Firstly, the data distribution is imbalanced, reflected in significant disparities in sample sizes between categories and uneven splits within categories. For instance, the \textbf{\textit{sn}} category has only 16.8k training samples, and the \textbf{\textit{mn}} category's test set is underrepresented. Secondly, reports within each category still contain considerable uninferable information due to the gap between the dataset and the clinical practice. After reconstruction, most categories in the MIMIC-RG4 dataset have larger sample sizes, and the splits for training, testing, and validation within each category are more balanced. Additionally, the amount of uninferable information in each category has been significantly reduced, while retaining the essential content.

\begin{table}[h]
	\centering
        \footnotesize
        \newcolumntype{C}{>{\centering\arraybackslash}X}%
		\begin{tabularx}{\columnwidth}{CCCCCC}
			\toprule
			Dataset	& split	& PC(\%) & PP(\%) & View(\%) & Comm(\%) \\
			\midrule
			\multirow{3}{*}{\makecell{MIMIC\\CXR}}  &Tr/84.3K&91.47&14.79&1.02&3.63 \\
			& Val/0.7K&92.84&15.81&1.49&3.38 \\
			& Ts/1.4K&93.06&16.26&1.33&7.22 \\
			\midrule
			\multirow{3}{*}{\makecell{MIMIC\\RG4}} &Tr/112.8K&77.36&0.28&	0.20&0.00\\
			& Val/0.9K&	78.55&	0.21&	0.11&	0.00 \\
            & Ts/2.0K&	85.29&	0.30&	0.49&	0.00 \\
			\bottomrule
		\end{tabularx}
		\caption{Percentage of reports with \textbf{\textit{sw}} setting, that encompass various categories of information. PC: Prior Comparison; PP: Prior Procedure; Comm: Communication; Tr: train; Ts: test.}
		\label{table1}
	\end{table}

 \begin{table}[h]
	\centering
        \footnotesize
        \newcolumntype{C}{>{\centering\arraybackslash}X}%
		\begin{tabularx}{\columnwidth}{CCCCCC}
			\toprule
			Dataset	& split	& PC(\%) & PP(\%) & View(\%) & Comm(\%) \\
			\midrule
			\multirow{3}{*}{\makecell{MIMIC\\CXR}}  &Tr/42.6K&27.09&8.18&35.13&2.89 \\
			& Val/0.3K&44.73&8.26&37.31&3.36 \\
			& Ts/0.1K	&74.80&	31.50&42.52&13.39 \\
			\midrule
			\multirow{3}{*}{\makecell{MIMIC\\RG4}} &Tr/91.3K&0.26&0.14&29.22&0.04\\
			& Val/0.7K&	0.43&	0.14&30.39&	0.00 \\
            & Ts/1.0K&	0.80&	0.20&33.17&	0.01 \\
			\bottomrule
		\end{tabularx}
		\caption{Percentage of reports with \textbf{\textit{mn}} setting, that encompass various categories of information. PC: Prior Comparison; PP: Prior Procedure; Comm: Communication; Tr: train; Ts: test.}
		\label{table1}
	\end{table}

 \begin{table}[h]
	\centering
        \footnotesize
        \newcolumntype{C}{>{\centering\arraybackslash}X}%
		\begin{tabularx}{\columnwidth}{CCCCCC}
			\toprule
			Dataset	& split	& PC(\%) & PP(\%) & View(\%) & Comm(\%) \\
			\midrule
			\multirow{3}{*}{\makecell{MIMIC\\CXR}}  &Tr/55.1K&74.77&15.64&28.64&2.47 \\
			& Val/0.4K&72.88&14.86&27.59&3.30 \\
			& Ts/0.9K	&88.34&	22.48&	36.69&	4.56 \\
			\midrule
			\multirow{3}{*}{\makecell{MIMIC\\RG4}} &Tr/47.7K&67.25&0.22&27.03&0.00\\
			& Val/0.4K&	64.69&	0.00&	25.34&	0.00 \\
            & Ts/0.8K&	92.87&	0.12&	33.57&	0.00 \\
			\bottomrule
		\end{tabularx}
		\caption{Percentage of reports with \textbf{\textit{mw}} setting, that encompass various categories of information. PC: Prior Comparison; PP: Prior Procedure; Comm: Communication; Tr: train; Ts: test.}
		\label{table1}
	\end{table}

\begin{table}[b!]
\centering
\footnotesize
\newcolumntype{C}{>{\centering\arraybackslash}X}%
\begin{tabularx}{\columnwidth}{CCCCCCCCCC}
    \toprule	 
    \multirow{2}{*}{\textbf{\( \mathbf{N_T} \).}} & \multirow{2}{*}{\textbf{ATF}}
    & \multicolumn{3}{c}{\textbf{CE Metrics}} & \multicolumn{3}{c}{\textbf{NLG Metrics}}&\multirow{2}{*}{\textbf{T}}\\ 
    \cmidrule(r){3-5} \cmidrule(r){6-8}
    & &P & R & F1 & B@1 & B@4 & R-L\\
    \midrule
    \multirow{2}{*}{\textbf{64}}& \xmark &\textbf{0.571}&0.580&0.575&0.463&0.194&0.391&21h\\
    & \cmark &0.570&\textbf{0.599}&\textbf{0.584}&\textbf{0.470}&\textbf{0.201}&\textbf{0.395}&\textbf{14h}\\
    \cmidrule(r){1-9}
    \multirow{2}{*}{\textbf{128}}& \xmark &0.563&0.592&0.577&0.469&0.197&0.391&25h\\
    & \cmark &\textbf{0.572}&\textbf{0.599}&\textbf{0.585}&\textbf{0.472}&\textbf{0.199}&\textbf{0.393}&\textbf{17h}\\
    \cmidrule(r){1-9}
    \multirow{2}{*}{\textbf{256}}& \xmark &\textbf{0.573}&0.591&0.582&\textbf{0.469}&\textbf{0.200}&0.395&35h\\
    & \cmark &0.572&\textbf{0.596}&\textbf{0.584}&0.465&0.199&\textbf{0.395}&\textbf{23h}\\
    \bottomrule
    \end{tabularx}
    \caption{Effects of ATF under different learnable tokens. \textbf{\( \mathbf{N_T} \).} represents the number of learnable tokens. \textbf{T} represents the training time per epoch.}
    \label{table1}
\end{table}

\subsection{Effects of ATF under Different Learnable Tokens}
The number of feature tokens is crucial for multimodal large language models. Generally, a higher number of feature tokens can provide the language model with richer information, leading to improved performance. To further investigate the feature extraction and compression capabilities of the ATF module under different feature token configurations, we select various numbers of learnable tokens and evaluate their performance in comparison with the interleaved input, as shown in Table 11. 
In all experiments, the TLW strategy is applied with the \(\lambda\) value set to 1.75.
We observe that with token numbers set to 64, 128, and 256, the model incorporating the ATF module demonstrate performance similar to that of interleaved input, indicating the scalability of ATF. Moreover, we find increasing the number of feature tokens does not yield significant improvements in this radiology report generation task.

\subsection{Trend of CE with Respect to \(\lambda\) in TLW}
To further explore how token-level loss weighting strategy influence the CE metrics, we evaluate the changes of F1 score (F1-5, F1-14) with respect to the hyperparameter \(\lambda\), as shown in Figure 5 and 6. We sequentially selecte six values for \(\lambda\), with the initial value corresponding to the absence of TLW strategy. The ATF module is employed in all experiments. 
We observe a steady increase in the model's F1 score with rising \(\lambda\), which plateaus after reaching a certain threshold. Additionally, we observe that excessively high values of lambda can slightly impair the BLEU-4 score, likely due to an overemphasis on key diagnostic tokens at the expense of overall report coherence.
These observations indicate that choosing an appropriate \(\lambda\) can effectively enhance the clinical efficacy without compromising NLG metrics.
Meanwhile, this phenomenon is evident in both the first and second stages, demonstrating a positive effect across different training stages.

\begin{figure}[h!]
	\centering
	\includegraphics[width=0.9\columnwidth]{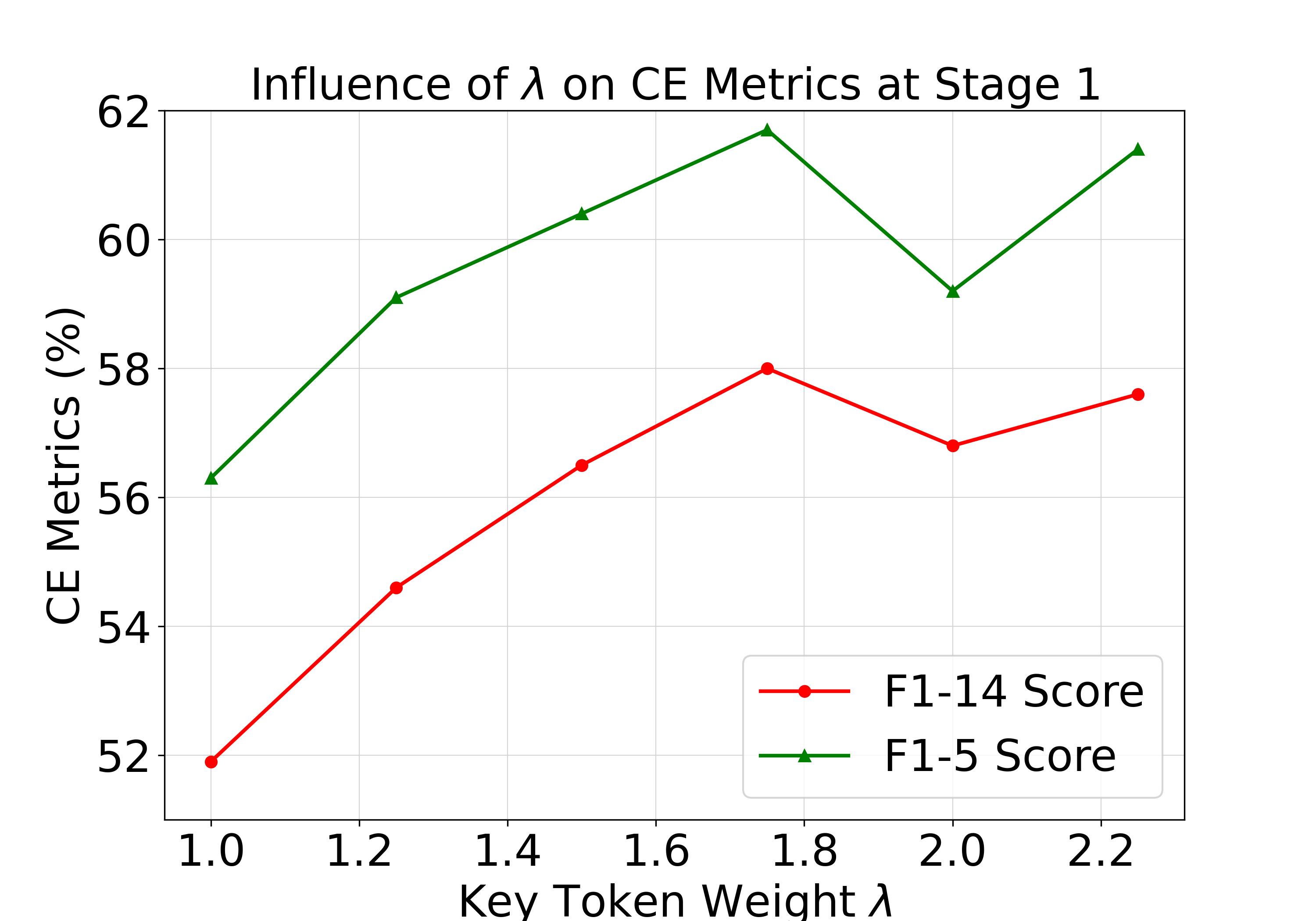}
	\caption{The influence of \(\lambda\) on CE metrics at stage 1.}
	\label{fig1}
\end{figure}

\begin{figure}[h!]
	\centering
	\includegraphics[width=0.9\columnwidth]{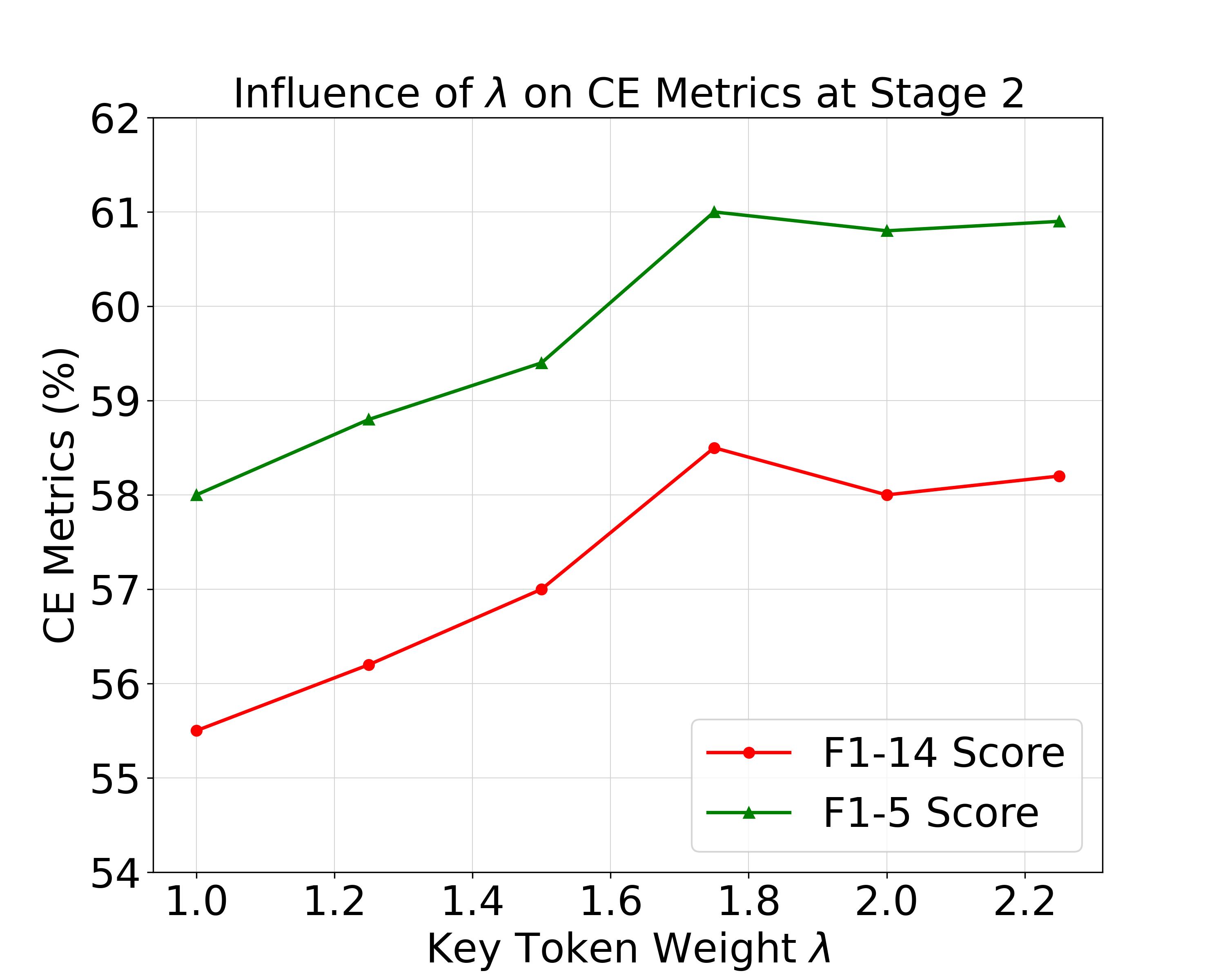} 
	\caption{The influence of \(\lambda\) on CE metrics at stage 2.}
	\label{fig1}
\end{figure}

 \subsection{LLM-RG4 Hyperparameters}
We set the maximum output length of Vicuna-7B v1.5 to be 150. The maximum truncation length during training is set to 100 and all reports are kept untruncated during testing. LoRA is implemented for all query and value projection matrices in the LLM, i.e., (\(W_q\), \(W_v\)). We set rank and $\alpha$ in LoRA to be 32 and 64 respectively. During mixed training, we upsample \textbf{\textit{sw}}, \textbf{\textit{mn}}, \textbf{\textit{mw}} category data to the size of \textbf{\textit{sn}}. When training in the specific setting, we extend the training epochs for \textbf{\textit{sw}}, \textbf{\textit{mn}}, \textbf{\textit{mw}} to ensure the number of iterations match or exceed that in the mixed training. Detailed hyperparameters used in inference, alignment stage and instruction stage are provided in Table 12, 13, and 14.
\begin{table}[h]
    \centering
    \footnotesize
    \begin{tabular}{ll} 
        \toprule
        \textbf{Hyperparameters} & \textbf{Values} \\ 
        \midrule
        Number of beams & 3 \\
        Maximum output length & 150 \\
        Minimum output length & 50 \\
        Repetition penalty & 2 \\
        Length penalty & 2 \\
        Temperature & 0 \\
        \bottomrule
    \end{tabular}
    \caption{Hyperparameters for LLM-RG4 inference.}
    \label{table6}
\end{table}

\begin{table}[h]
    \centering
    \footnotesize
    \begin{tabular}{ll} 
        \toprule
        \textbf{Hyperparameters} & \textbf{Values} \\ 
        \midrule
        Optimizer & AdamW \\
        Weight decay & 0.01 \\
        Betas &  [0.9, 0.999] \\
        Learning rate & $3 \times 10^{-4}$ \\
        Warmup rates & 0.1 \\
        Number of workers & 12 \\
        Type of GPU & NVIDIA A800 \\
        Number of GPU & 1 \\
        Accumulate gradient batches & 2 \\
        Batch size per GPU (total) & 24(48) \\
        Training precision & bfloat16\\
        Epochs & 2\\
        \bottomrule
    \end{tabular}
    \caption{Hyperparameters for the alignment stage.}
    \label{table7}
\end{table}

\begin{table}[h]
    \centering
    \footnotesize
    \begin{tabular}{ll} 
        \toprule
        \textbf{Hyperparameters} & \textbf{Values} \\ 
        \midrule
        Optimizer & AdamW \\
        Weight decay & 0.01 \\
        Betas &  [0.9, 0.999] \\
        Learning rate & $3 \times 10^{-4}$ \\
        Warmup rates & 0.1 \\
        Number of workers & 12 \\
        Type of GPU & NVIDIA A800 \\
        Number of GPU & 1 \\
        Accumulate gradient batches & 2 \\
        Batch size per GPU (total) & 16(32) \\
        Training precision & bfloat16\\
        Epochs & 2\\
        \bottomrule
    \end{tabular}
    \caption{Hyperparameters for the instruction-tuning stage.}
    \label{table8}
\end{table}

\begin{table*}[t]
    \centering
    \footnotesize
    \begin{tabular}{p{1.0\textwidth}}
        \toprule
        \textbf{Prior Comparison Judgement Rule:} \\
        \textbf{System Message:} Be an X-ray radiology assistant \\
        \textbf{User Message:} You will be given a chest X-ray report. Please check if there are any description or mention of comparisons or references to previous xray inspections in this report. For example, expression like previous discoveries have changed, new discoveries have emerged, some things have been renoticed, some things are unchanged and treatment equipment has changed often implies a comparison. The report may include words such as new, stable, improv, resol, disappear, prior, stable, previous, again, remain, remov, similar, earlier, decreas, recurr, redemonstrate, etc. Please note that the expression of the word /stable/ is stable compared to previous reports and also belongs to the category of comparison. Please pay attention to the hidden comparisons in the expression. If it contains any comparison, please answer yes, else answer no. Just answer yes or no. \\
        \textbf{Report:} \dots\\
        \textbf{Judgement Result:} \dots\\
        \\
        \textbf{Prior Procedure Judgement Rule:} \\
        \textbf{System Message:} Be an X-ray radiology assistant \\
        \textbf{User Message:} You will be given a chest X-ray report. Please check if the report includes one of the following three contents.  \\
        1.Please check if there is any reference to or mention of or comparison with CT or MRI or PET examination done before, rather than comparison with prior radiograph. \\
        2.Please check if there is an explicitly description of the patient's postoperative status included, rather than guessing. \\
        3.Please check if there is any explicitly removal of previous treatment devices such as tubes, clips, Port-A-Cath and so on, rather than implication. \\
        If report includes any of them, please answer yes, else answer no. Just answer yes or no. \\
        \textbf{Report:} \dots \\
        \textbf{Judgement Result:} \dots\\
        \\
        \textbf{View Description Judgement Rule:} \\
        \textbf{System Message:} Be an X-ray radiology assistant \\
        \textbf{User Message:} You will be given a chest X-ray report. Please check if it is written using a single AP or PA image as input. Specifically, please check if the report includes one of the following two contents. \\
        1.Please check if there is any explicitly mention about what is seen in the lateral view image. \\
        2.Please check if there is direct statement indicating that both frontal (AP or PA) and lateral view image were provided to evaluate or compare. \\
        Besides these two condition, please note that the mention of recommending lateral view to confirm does not fall under the above two condition and a descriptive term for the location does not necessarily imply that a separate lateral view image was examined. If the report meets any condition, it indicates that it was written using both frontal and lateral positions as inputs. Please answer yes, otherwise answer no. Just answer yes or no.' \\
        \textbf{Report:} \dots \\
        \textbf{Judgement Result:} \dots\\
        \\
        \textbf{Communication Description Judgement Rule:} \\
        \textbf{System Message:} Be an X-ray radiology assistant \\
        \textbf{User Message:} You will be given a chest X-ray report. Please check if there is any mention about communication between medical professionals, such as between doctors or nurses. Please check if there is any description of transfering inspection results. Please note that suggestions are not part of communication. If is contains, please answer yes, else answer no. Just answer yes or no. \\
        \textbf{Report:} \dots \\
        \textbf{Judgement Result:} \dots\\
        \bottomrule
    \end{tabular}
    \caption{Prompts for report judgement.}
    \label{tab:judgement_prompts}
\end{table*}

\begin{table*}[t]
    \centering
    \footnotesize
    \begin{tabular}{p{1.0\textwidth}}
        \toprule
        \textbf{Prior Comparison Reconstruction Rule:} \\
        \textbf{System Message:} Be an X-ray radiology assistant \\
        \textbf{User Message:} Prior Comparison Judgement Instruction + Report \\
        \textbf{Assistant Message:} Judgement result \\
        \textbf{User Message:} "Please rewrite the report to remove the above mentioned information while maintaining the original meaning as much as possible. Here are some rewriting principles. Please rewrite them according to the corresponding rules based on the reasons. \\
        1. For sentences that mention comparison with a certain check, they often contain `\_'. Delete the sentence that includes the comparison behavior, but retain the result; \\
        2. For adjectives related to time, such as new and old, delete them directly. For adverbs related to time and program, such as again and so, delete them directly; \\
        3. If there is no change, slight improvement, or deterioration in the description of previous positive symptoms that still exist, rewrite it to directly state this symptom;  \\
        4. For sentences describing a symptom that has disappeared, rewrite them as negative and mention it.  \\
        Please only give me the rewritten report, and if there is a situation where the above rules cannot be rewritten, please output: `attention! a certain expression exceeds the current rule range.'  \\
        \textbf{Rewritten Report:} \dots\\
        \\
        \textbf{Prior Procedure Reconstruction Rule:} \\
        \textbf{System Message:} Be an X-ray radiology assistant \\
        \textbf{User Message:} Prior Procedure Judgement Instruction + Report \\
        \textbf{Assistant Message:} Judgement result \\
        \textbf{User Message:} Please rewrite the report to remove the above mentioned information while maintaining the original meaning as much as possible. Here are some rewriting principles. Please rewrite them according to the corresponding rules based on the reasons. \\
        1. For sentences that mention the patient's postoperative state, delete the description of what postoperative state they are in;  \\
        2. For examination results obtained using other non X-ray examination methods, such as CT, MRI, PET, these examination results shall be directly deleted;  \\
        3. For sentences describing the removal of previous treatment devices such as pacemakers, clips, etc., delete them directly.  \\
        Please only give me the rewritten report, and if there is a situation where the above rules cannot be applied, please answer: `attention! a certain expression exceeds the current rule range.' \\
        \textbf{Rewritten Report:} \dots\\
        \\
        \textbf{View Description Reconstruction Rule:} \\
        \textbf{System Message:} Be an X-ray radiology assistant \\
        \textbf{User Message:} View Description Judgement Instruction + Report \\
        \textbf{Assistant Message:} Judgement result \\
        \textbf{User Message:} Please rewrite the report to remove the above mentioned information while maintaining the original meaning as much as possible. Here are some rewriting principles. Please rewrite them according to the corresponding rules based on the reasons. \\
        1. For expressions provided for different views, such as AP and later views are provided, remove the expression directly while retaining the inspection results. \\
        2. For statements that clearly state what symptoms are seen or confirmed from the lateral view, remove the statement and replace it with the statement that `the lateral view is recommended to assist in diagnosis.' \\
        3. For conclusions obtained from both frontal and lateral views, delete the description of the view and retain the conclusion.  \\
        Please only give me the rewritten report, and if there is a situation where the above rules cannot be applied, please answer: `attention! a certain expression exceeds the current rule range.' \\
        \textbf{Rewritten Report:} \dots\\
        \\
        \textbf{Communication Description Reconstruction Rule:} \\
        \textbf{System Message:} Be an X-ray radiology assistant \\
        \textbf{User Message:} Communication Description Judgement Instruction + Report \\
        \textbf{Assistant Message:} Judgement result \\
        \textbf{User Message:} Please rewrite the report to remove the above mentioned information while maintaining the original meaning as much as possible. Here are some rewriting principles. Please rewrite them according to the corresponding rules based on the reasons. \\
        1. If a sentence describes communication with doctors and nurses, delete that sentence directly. \\
        Please only give me the rewritten report, and if there is a situation where the above rules cannot be applied, please answer: `attention! a certain expression exceeds the current rule range.' \\
        \textbf{Rewritten Report:} \dots\\
        \midrule
        \textbf{Iterative Modification Rules for Scenario \(X\):}\\
        \textbf{System Message:} Be an X-ray radiology assistant \\
        \textbf{User Message:} Judgement Instruction \(X\) + Report \\
        \textbf{Assistant Message:} Judgement result 1\\
        \textbf{User Message:} Reconstruction Instruction \(X\)\\
        \textbf{Assistant Message:} Rewritten report 1\\
        \textbf{User Message:} Judgement Instruction \(X\) + Rewritten Report 1\\
        \textbf{Assistant Message:} Judgement result 2\\
        \textbf{User Message:} Reconstruction Instruction \(X\)\\
        \textbf{Rewritten Report2:} \dots\\
        \bottomrule
    \end{tabular}
    \caption{Prompts for report reconstruction.}
    \label{tab:rewriting_prompts}
\end{table*}

\begin{figure*}[t]
	\centering
	\includegraphics[width=0.9\textwidth]{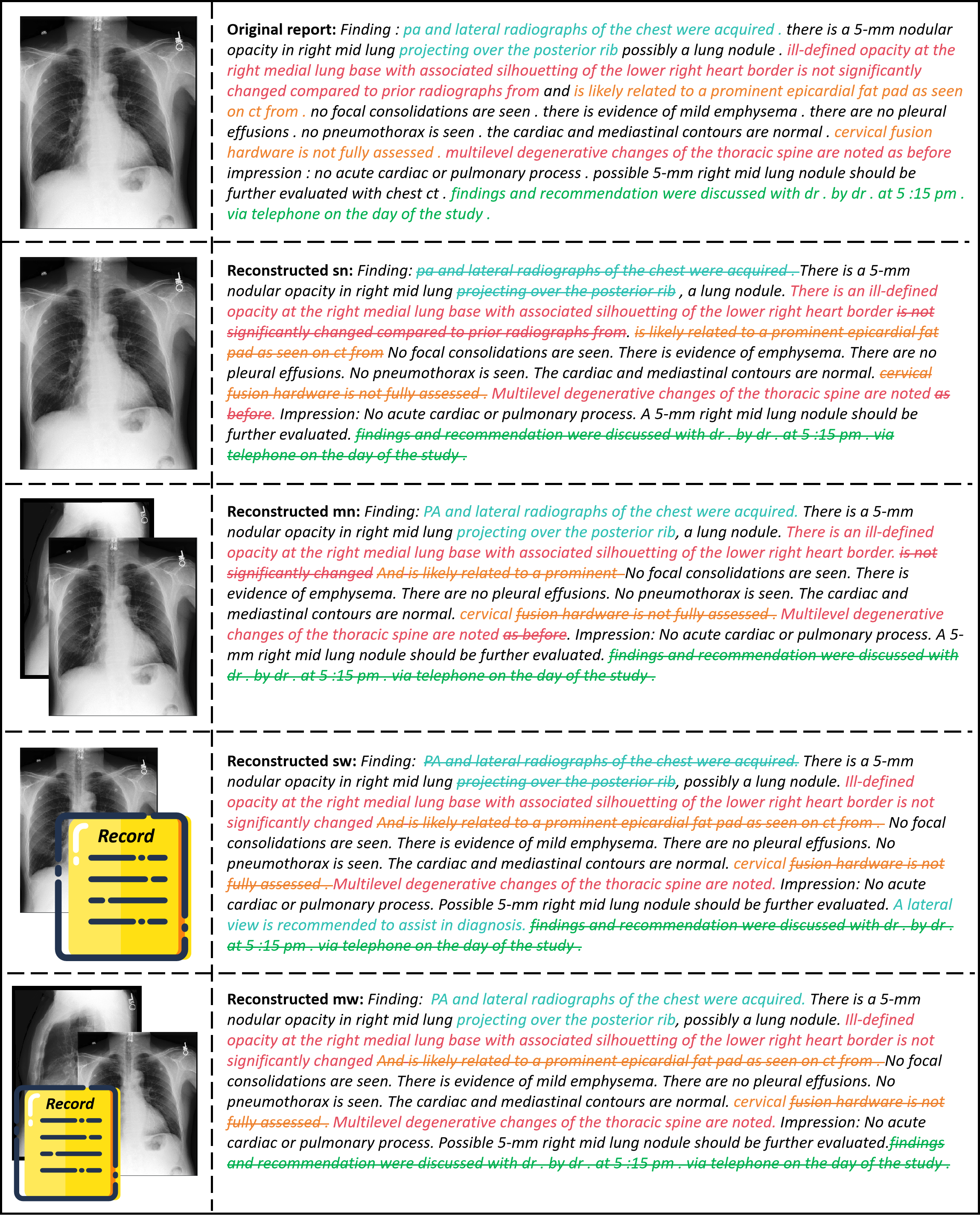} 
	\caption{A qualitative illustration of the reconstructed reports. We use strikethrough to indicate the differences between the original reports and reconstructed reports. Different category descriptions are highlighted using the same color, red: prior comparison; orange: prior procedure; blue: view description; green: communication. The reconstructed reports show close alignment with the input while minimizing information loss.}
	\label{fig3}
\end{figure*}

\end{document}